\documentclass[10pt,twocolumn,letterpaper]{article}

\usepackage[pagenumbers]{wacv} %

\usepackage{algorithm,algpseudocode}
\usepackage{multirow}
\usepackage{listings}
\usepackage{caption}
\usepackage{times}
\usepackage{epsfig}
\usepackage{graphicx}
\usepackage{amsmath}
\usepackage{amssymb}
\usepackage{subcaption}
\usepackage{tocloft}
\usepackage[title]{appendix}

\usepackage[table, dvipsnames]{xcolor}
\newcommand{\red}[1]{{\color{red}#1}}

\newcommand{\hlb}[1]{\cellcolor{orange!30}\textbf{#1}}
\newcommand{\hli}[1]{\cellcolor{orange!30}\textit{#1}}

\usepackage{amsmath}
\DeclareMathOperator*{\argmax}{arg\,max_{k}}

\definecolor{cvprblue}{rgb}{0.21,0.49,0.74}
\definecolor{boxgreen}{rgb}{0.21,0.49,0.13}
\definecolor{boxblue}{rgb}{0.07,0.0,0.96}

\newcommand{\method}{PDV\xspace}

\definecolor{cvprblue}{rgb}{0.21,0.49,0.74}
\definecolor{boxgreen}{rgb}{0.21,0.49,0.13}
\definecolor{boxblue}{rgb}{0.07,0.0,0.96}

\usepackage[pagebackref,breaklinks,colorlinks,citecolor=cvprblue]{hyperref}
\usepackage{xspace}

\title{PDV: Prompt Directional Vectors for Zero-shot Composed Image Retrieval}

\author{Osman Tursun\textsuperscript{1}, Sinan Kalkan\textsuperscript{2}, Simon Denman\textsuperscript{1} and Clinton Fookes\textsuperscript{1}\\
	\textsuperscript{1}Queensland University of Technology\\
	\textsuperscript{2}Middle East Technical University\\
	{\tt\small \{osman.tursun,s.denman,c.fookes\}@qut.edu.au,skalkan@metu.edu.tr}
}

\begin{document}

\maketitle

\begin{abstract}
Zero-shot Composed Image Retrieval (ZS-CIR) enables image search using a reference image and a text prompt without requiring specialized text-image composition networks trained on large-scale paired data. However, current ZS-CIR approaches suffer from three critical limitations in their reliance on composed text embeddings: static query embedding representations, insufficient utilization of image embeddings, and suboptimal performance when fusing text and image embeddings. To address these challenges, we introduce the \textbf{Prompt Directional Vector (PDV)}, a simple yet effective training-free enhancement that captures semantic modifications induced by user prompts. PDV enables three key improvements: (1) Dynamic composed text embeddings where prompt adjustments are controllable via a scaling factor, (2) composed image embeddings through semantic transfer from text prompts to image features, and (3) weighted fusion of composed text and image embeddings that enhances retrieval by balancing visual and semantic similarity. Our approach serves as a plug-and-play enhancement for existing ZS-CIR methods with minimal computational overhead. Extensive experiments across multiple benchmarks demonstrate that PDV consistently improves retrieval performance when integrated with state-of-the-art ZS-CIR approaches, particularly for methods that generate accurate compositional embeddings. The code will be released upon publication.
\end{abstract}

\section{Introduction}
\label{sec:intro}

Composed Image Retrieval (CIR) involves searching for images using a combination of a reference image and a prompt that describes how the target image should differ from the reference \cite{ vo2019composing, baldrati2022effective,baldrati2023zero,saito2023pic2word}. Compared to traditional content-based image retrieval (CBIR) systems, CIR offers increased flexibility and precision by allowing users to articulate complex, multi-modal queries that combine visual and semantic information \cite{saito2023pic2word,karthik2024visionbylanguage,cohen2022my}. %

\begin{figure*}[!ht]
	\centering
	\includegraphics[width=1\linewidth]{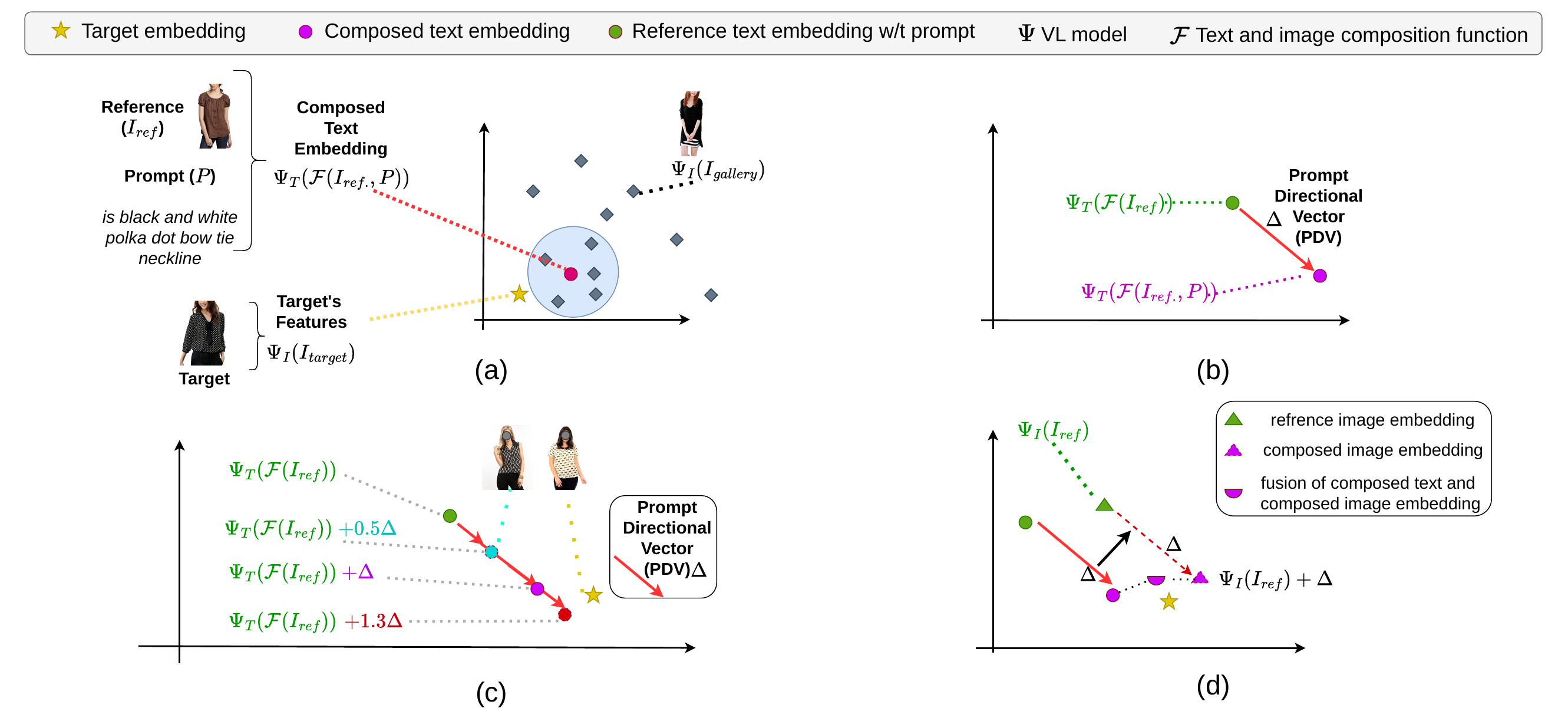}
	\caption{Overview of Prompt Directional Vector (PDV) for Zero-Shot Composed Image Retrieval (ZS-CIR). (a) Standard ZS-CIR pipeline. (b) PDV calculation process. (c) Dynamic text embedding composition using PDV. (d) Fusion of composed text and image embeddings: PDV-modified composed image embedding combined with composed text embedding.}
	\label{fig:motiv}
\end{figure*}

The core challenge in CIR lies in effectively integrating information from two distinct modalities: image and text. With the rapid progress in vision and language models (VLMs), CIR has attracted significant attention in the computer vision community \cite{liu2021image,baldrati2022effective,karthik2024visionbylanguage,saito2023pic2word,baldrati2023zero}. Early approaches to CIR were primarily supervised in nature \cite{kim2021dual,anwaar2021compositional,vo2019composing,chen2020learning,wu2021fashion,lee2021cosmo}. However, as highlighted by Saito et al. \cite{saito2023pic2word}, the labeling cost for supervised datasets in this domain is prohibitively high, prompting researchers to explore more efficient alternatives, namely zero-shot composed image retrieval (ZS-CIR). In this work, we provide a simple and training-free approach to improve the controllability and accuracy of existing ZS-CIR approaches.

ZS-CIR leverages VLMs, denoted by $\Psi$, which operate through a dual-pathway architecture. The first pathway consists of a vision branch, $\Psi_I$, that extracts feature representations from target images, $I_{target}$. The second pathway employs a language branch, $\Psi_T$, that processes a textual composition of reference images, $I_{ref}$, and user-provided text prompts, $P$. This composition, represented by $\mathcal{F}(I_{ref}, P)$, can be achieved through two primary methods: (1) \textit{Caption Generation}, where a caption is generated for the reference image using a VLM, and this caption is merged with the text-prompt using Large Language Models (LLMs), as demonstrated in CIReVL \cite{karthik2024visionbylanguage}; or (2) \textit{Pseudo Tokenization}, which uses CLIP's \cite{radford2021learning} visual branch to process $I_{ref}$ and a mapping network (consisting of a lightweight multi-layer perceptron) to tokenise the image, as demonstrated in Pic2Word \cite{saito2023pic2word}. The resulting $\mathcal{F}(I_{ref},P)$ is a textual query representation that encompasses both the provided visual and text information, and facilitates zero-shot retrieval.
The aforementioned pipeline is illustrated in Figure \ref{fig:motiv}\red{a}. 

We identify three major gaps in the literature, despite the promising results \cite{saito2023pic2word,karthik2024visionbylanguage,gu2024lincir,baldrati2023zero}:

\noindent{\bf{Gap 1: Inefficient Iterative Search Due to Static Composed Embeddings}} Existing CIR studies primarily emphasize initial retrieval success but neglect users' need to iteratively refine and steer results when initial searches fail. This limitation stems mainly from the static nature of composed text embeddings. Existing ZS-CIR approaches are unable to directly tune the composed text embedding, $\Psi_{T}(\mathcal{F}(I_{ref},P))$, to bring it closer to the target image embedding, $\Psi_{T}(\mathcal{F}(I_{ref},P))$. With existing ZS-CIR approaches users need to refine their prompts to iteratively improve retrieval results. However, this approach incurs additional manual effort for prompt construction and computational overhead from expensive feature extraction.

\noindent{\bf{Gap 2: Underutilisation of Reference Image Embedding}}. Current methods generally do not utilize the embedding $\Psi_{I}(I_{ref})$ of the reference image directly for retrieval; instead, $\Psi_{I}(I_{ref})$ is used solely for composition. This omission stems from consistently poor retrieval performance (refer to Image-only results in Table \ref{tab:circo_cirr_results_pdv-I}) when incorporating these embeddings, as documented in multiple studies \cite{saito2023pic2word,karthik2024visionbylanguage,baldrati2023zero}.

\noindent{\bf{Gap 3: Suboptimal Performance of Image-Text Embedding Fusion}}. While the fusion of image and text embeddings outperforms single-modality approaches (image-only or text-only) \cite{saito2023pic2word,karthik2024visionbylanguage,baldrati2023zero}, it still underperforms compared to composed text embeddings (refer to ``Image+Text" results in Tables \red{S5, S6} in the supplementary.).

\noindent{\bf{Promp Directional Vector (PDV): a Plug-and-Play Solution}}. We propose the \emph{Prompt Directional Vector} (PDV) as a straightforward, training-free approach to address the aforementioned gaps. Denoted by $\Delta_{PDV}$, the PDV represents the residual vector between two text embeddings: the composed text embedding $\Psi_{T}(\mathcal{F}(I_{ref},P))$ and the reference image text embedding $\Psi_{T}(\mathcal{F}(I_{ref}))$. The latter is equivalent to $\Psi_{T}(\mathcal{F}(I_{ref}, P_{Empty}))$, where $P_{Empty}$ represents an empty input string, corresponding to the unprompted baseline. As illustrated in Figure \ref{fig:motiv}\red{b} and shown via a red arrow, this PDV captures the semantic modification induced by the prompt. In the following, we summarize how the PDV effectively addresses the three aforementioned challenges.

\noindent{\bf{\method: Addressing Gap 1}}. To change the static nature of the composed text embedding and increase flexibility and utility for users, we generalise the synthesis of the composed text embeddings $\Psi_{T}(\mathcal{F}(I_{ref}, P))$. We interpret $\Psi_{T}(\mathcal{F}(I_{ref}, P)$ as a shift from the reference image text embedding without the prompt, $\Psi_{T}(\mathcal{F}(I))$, by a vector $\Delta_{PDV}$. Under this formulation, the baseline ZS-CIR approach can be viewed as a special case where $\Psi_{T}(\mathcal{F}(I_{ref}, P)) = \Psi_{T}(\mathcal{F}(I_{ref}, P)) + \alpha\Delta_{PDV}$ with $\alpha=1$. We hypothesize that when $\Delta_{PDV}$ captures the desired modifications, but not their precise magnitude (particularly with less descriptive prompts), adjusting $\alpha$ can enhance retrieval performance and controllability. As demonstrated in Figure \ref{fig:motiv}\red{c}, increasing $\alpha$ to 1.3 produces results more closely aligned with the target compared to the default $\alpha=1$.

\noindent{\bf{\method: Addressing Gap 2}}. Although image embeddings $\Psi_{I}(I_{ref})$ contain valuable visual content regarding the reference image, they lack prompt-specific semantic information, leading to poor performance when used in ZS-CIR. By leveraging the shared semantic space learned by Vision-Language models, we can transfer prompt semantics to the image embedding by adding the Prompt Vector $\Delta$, obtaining $\Psi_{I}(I_{ref}) + \alpha\Delta_{PDV}$, as illustrated in Figure \ref{fig:motiv}\red{d}. We denote this augmented representation as the \textit{composed image embedding}. Similar to the dynamic composed text embedding, this representation can be adjusted through a scaling factor, $\alpha$ to offer controllability to enhance retrieval.

\noindent{\bf{\method: Addressing Gap 3}}. 
Lastly, several studies demonstrate that the direct fusion of image and text embeddings outperforms using either input feautre (image or text-prompt) alone \cite{saito2023pic2word,karthik2024visionbylanguage,baldrati2023zero}. However, this fusion approach still underperforms compared to using the composed text embeddings. This performance gap exists because prompt embeddings are significantly changed by incorporating context from the reference image. Specifically, $\Delta_{PDV}$ is not equivalent to $\Psi_{T}(P)$. To address this, we propose fusing the composed text and composed image embeddings, as illustrated in Figure \ref{fig:motiv}\red{d}. Through varying the fusion weight factor $\beta$, we can dynamically control the balance between visual similarity to the reference image and semantic alignment with the prompt without needing to craft new prompts, or modify reference images. Lower $\beta$ values prioritize visual fidelity, while higher values emphasize semantic modifications specified in the prompt. The previous work, CompoDiff \cite{gu2023compodiff}, also demonstrates similar controllability through adjusting visual and semantic weights. However, their method cannot be applied to other approaches as ours can, as our method is an add-on approach while theirs is specifically tailored to their own framework.

PDV serves as a plug-and-play enhancement for most ZS-CIR approaches, offering a simple and training-free solution. The computational overhead is minimal, requiring only the calculation of text and image embeddings from the reference image. We evaluate PDV by integrating it with four distinct ZS-CIR methods across various CIR benchmarks. Our experimental results demonstrate that all three use cases of PDV consistently improve upon baseline approaches, particularly when the baseline method already generates accurate compositional embeddings.

\noindent{\textbf{Contributions}}. Our main contributions are as follows:
(i) We introduce the Prompt Directional Vector (PDV), a simple and training-free enhancement that overcomes limitations of current Zero-Shot CIR methods. (ii) We propose three novel applications of PDV: (1) dynamic composed text embedding synthesis through PDV scaling, which offers enhanced control over retrieval results without tedious prompt modification; (2) composed image embedding synthesis via semantic transfer of prompts to visual features through PDV addition, which prioritizes visual similarity; and (3) effective fusion of composed text and image embeddings, which improves overall performance and enables controllable balancing of visual and semantic similarity. (iii) Through extensive experiments on multiple benchmarks with four ZS-CIR methods, we demonstrate that PDV consistently improves retrieval performance with minimal computational overhead.

\section{Related Work}
\label{sec:lit}
\textbf{Vision-Language (VL) models} have revolutionized computer vision by effectively bridging visual and textual modalities. The emergence of powerful models such as CLIP \cite{radford2021learning}, ALIGN \cite{jia2021scaling}, and Florence \cite{yuan2021florence} has enabled remarkable advances in multi-modal understanding. Trained on large-scale image-text pairs through contrastive learning, these models learn rich visual-semantic representations that generalize across domains and tasks.
Building upon these advances, Composed Image Retrieval (CIR) has shown significant progress \cite{baldrati2022effective,saito2023pic2word,karthik2024visionbylanguage}. Early approaches leveraged VL models and either trained a combiner network to compose text and image features \cite{baldrati2022effective}, or fine-tuned a text encoder \cite{baldrati2022conditioned} to extract task-specific text features. However, these methods required expensive domain-specific triplets (reference image, modified image, and text description) that must be manually verified. Recent work has explored alternative approaches to reduce the data collection burden, such as using synthetic triplets \cite{gu2023compodiff} or mining triplets from large-scale image-text datasets \cite{liu2023zeroshot}. However, these methods still incur significant computational costs during training.

\noindent{{\bf Zero-shot CIR with Text Inversion}} Recent research has focused on zero-shot approaches to address these challenges. Many methods adopt \textit{text inversion}, a technique initially proposed for personalized image generation \cite{gal2022textual,ruiz2023dreambooth}, which maps images to pseudo-tokens or words. Pic2Word \cite{saito2023pic2word} introduced a self-supervised text inversion network trained with cyclic contrastive loss, though it requires a large-scale image dataset. SEARLE \cite{baldrati2023zero} reduces the cost of training Pic2Word and improves the efficiency of the text inversion network. KEDs \cite{suo2024knowledge} implicitly models the attributes of the reference images by incorporating a database; thus, tokens obtained through inversion include attributes such as color, object number and layout. To further improve scalability, LinCIR \cite{gu2024lincir} proposed a language-only approach that reduces training costs and increases scalability. Most recently, CIReVL \cite{karthik2024visionbylanguage} introduced a more direct approach that leverages image captioning models to generate natural language descriptions of reference images, which are then combined with text that specifies desired modifications to form queries. Subsequently proposed methods, such as LDRE \cite{yang2024ldre} and SEIZE \cite{yang2024semantic}, leverage multiple captions over a single caption to increase the diversity and also take the semantic increment during the composition into consideration.

\noindent{{\bf Composition with a Residual}} 
In contrast to ZS-CIR, early supervised CIR approaches learned prompt-induced modifications by training on labelled triplet data (reference image, prompt, and target image). Vo et al. \cite{vo2019composing} pioneered this approach by introducing a residual learning module based on an LSTM network. Subsequently, several methods \cite{chen2020learning,wu2021fashion,yu2020curlingnet} adopted similar residual learning strategies for text-image composition. Baldrati et al. \cite{baldrati2022conditioned} further advanced this approach by fine-tuning CLIP's text encoder to learn residual embeddings.
While these prior works explored residual-based approaches, they all relied on supervised training. In contrast, our proposed PDV achieves similar capabilities by directly leveraging pre-trained VL models, eliminating the need for task-specific training.

\section{Methodology}
\label{sec:meth}

\subsection{Baseline ZS-CIR Framework}
Composed Image Retrieval (CIR) enables users to search for target images $I_{target}$ by providing a reference image, $I_{ref}$, and a text prompt, $P$, describing desired modifications. Zero-shot composed image retrieval (ZS-CIR) leverages Vision-Language (VL) models, $\Psi$, such as CLIP \cite{radford2021learning}, whose vision branch, $\Psi_{I}$, and text branch, $\Psi_{T}$, are trained to learn a shared embedding space where semantically similar image and text pairs are mapped close to each other. In this framework, as show in Figure \ref{fig:motiv}\red{a}, target images are encoded using the vision branch, $\Phi_{I}$, while the query is composed by processing both $I_{ref}$ and $P$ through the text branch $\Psi_{T}$, as composition operations are more naturally handled in the text modality. 

Recent ZS-CIR approaches generate the composed text embedding from $I_{ref}$ and $P$ using one of two methods: direct image captioning (CIReVL, LDRE and SEIZE) or pseudo tokenization (Pic2Word, LinCIR, SEARLE and KEDs). We denote this composition process as $\mathcal{F}$, resulting in a composed text embedding $\Psi_T(\mathcal{F}(I_{ref}, P))$.

In an ideal ZS-CIR scenario, the target image $I_{target}$ should appear within the top-k results retrieved from the gallery $\mathcal{D}$. This retrieval is formalized as:
\begin{equation}
\label{eq:zscir}
\mathbb{I}_{top-k} = \argmax_{I \in \mathcal{D}} \frac{\Psi_T(\mathcal{F}(I_{ref}, P))^T \cdot \Psi_I(I)}{\lVert\Psi_T(\mathcal{F}(I_{ref}, P))\rVert \cdot \lVert\Psi_I(I)\rVert}.
\end{equation}
If $I_{target} \notin \mathbb{I}_{top-k}$, the user must reformulate the prompt and repeat the feature extraction process to obtain alternative retrieval results, incurring time and computational resource costs. Notably, as shown in Eq. \ref{eq:zscir}, only the composed feature embedding $\Psi_T(\mathcal{F}(I_{ref}, P))$ directly influences the computation of $\mathbb{I}_{top-k}$. Although the gallery images are represented by their image embeddings, the image embedding of the reference image $\Psi_I(I_{ref})$ do not contribute to the retrieval process.

\subsection{Our Approach: Prompt Directional Vector}

Rather than simply employing the composed embedding alone, $\Psi_T(\mathcal{F}(I_{ref}, P))$, as depicted in Figure \ref{fig:motiv}\red{b},
we propose a generalized formulation of composed text embeddings by considering the embedding modification direction, $\delta_{PDV}$, which is derived from the difference between the provided prompt, $P$, and the reference image, $I_{ref}$. Formally, we define $\delta_{PDV}$ as,
\begin{equation}
\Delta_{PDV} = \Psi_T(\mathcal{F}(I_{ref}, P)) - \Psi_T(\mathcal{F}(I_{ref})).
\label{eq:pdv}
\end{equation}
We then form the composed text embedding as follows,
\begin{equation}
	\Psi_T(\mathcal{F}(I_{ref}, P)) = \Psi_T(\mathcal{F}(I_{ref})) + \alpha_T\Delta_{PDV}, \label{eqn:text_embedding}
\end{equation}
{\noindent}where $\alpha$ controls the movement along the prompt vector $\Delta_{PDV}$ and $\Psi_T(\mathcal{F}(I_{ref}))$ is the original text embedding.

\subsection{Strategies for Using PDV}

We explore three strategies for using $\Delta_{PDV}$:

\textbf{(1) Prompt Directional Vector for Text (PDV-T)}, which enhances controllability in ZS-CIR. While baseline ZS-CIR approaches represent a special case where $\alpha=1$, varying $\alpha$ provides users with additional control over the retrieval process (refer to Figure \ref{fig:motiv}\red{c}). Setting $\alpha>1$ amplifies the modification specified by the prompt, while $\alpha<1$ reduces its effect. This approach offers a more efficient alternative to modifying the prompt directly, as it requires neither new feature extraction nor prompt reformulation. Note that we use the notation $\Phi_{PDV-T}$ to represent the composed text embedding.

\begin{figure*}[!tb]
	\centering
	\begin{subfigure}[b]{0.495\linewidth}
		\centering
		\includegraphics[width=\linewidth]{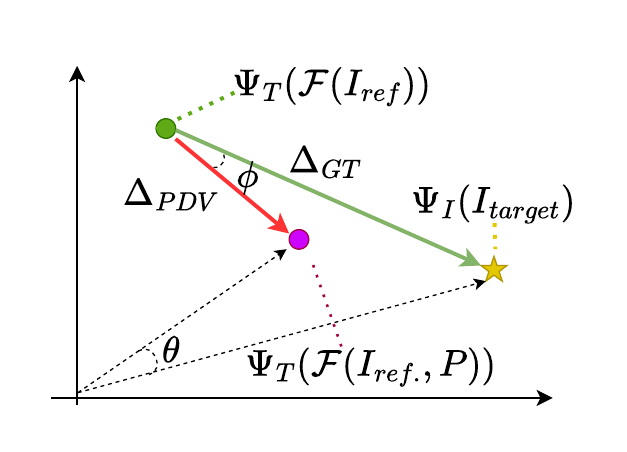}
		\caption{Visualisation of $\theta$ and $\phi$.}
		\label{fig:theta-a}
	\end{subfigure}
	\begin{subfigure}[b]{0.495\linewidth}
		\centering
		\includegraphics[width=\linewidth]{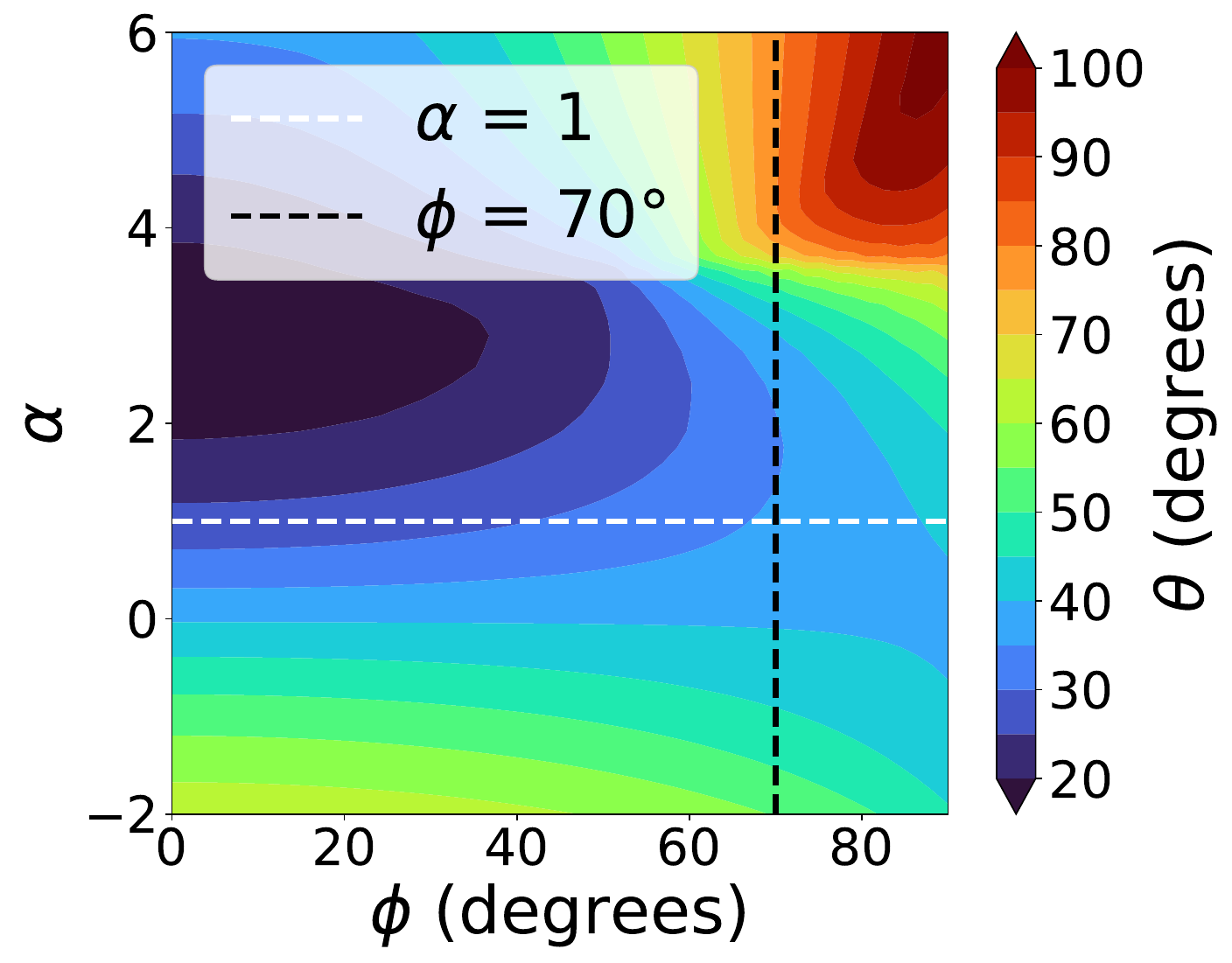}
		\caption{$\theta$ vs. $\alpha$ and $\phi$}
		\label{fig:theta-b}
	\end{subfigure}
	\caption{A visualisation of how scaling $\alpha$ affects the $
    \theta$ angle between the composed embedding and the target embedding. }
	\label{fig:theta}
\end{figure*}

\noindent{\bf When is PDV-T effective?} In CIR, better retrieval performance is directly correlated with a smaller angle, $\theta$,  between the target embedding vector $\Psi_{I}(I_{target})$ and the composed embedding vector $\Psi_T(\mathcal{F}(I_{ref}, P))$ (refer to Figure~\ref{fig:theta-a}). When $\phi$, the angle between the calculated prompt directional vector $\Delta_{\mathrm{PDV}}$ and the ground truth prompt directional vector $\Delta_{\mathrm{GT}}$, is small, adjusting the parameter $\alpha$ can effectively reduce $\theta$. To demonstrate this relationship, we conducted a simulation based on a common real-world scenario where the magnitude of $\Delta_{\mathrm{PDV}}$ is smaller than $\Delta_{\mathrm{GT}}$ due to users writing less descriptive prompts (see Figure~\ref{fig:theta-a}). Based on simulation results (Figure~\ref{fig:theta-b}), when $\phi$ is less than 70 degrees, increasing $\alpha$ up to 3 reduces $\theta$. However, when $\phi$ is between 70 and 90 degrees, decreasing $\alpha$ is more effective in reducing $\theta$. These findings indicate that PDV's performance is highly dependent on the accuracy of the baseline CIR method. With a strong baseline system, adjusting $\alpha$ becomes a straightforward and effective way to tune CIR results.

\begin{figure}[!tbh]
	\vspace*{-0.4cm}
	\centerline{
		\includegraphics[width=1.1\linewidth]{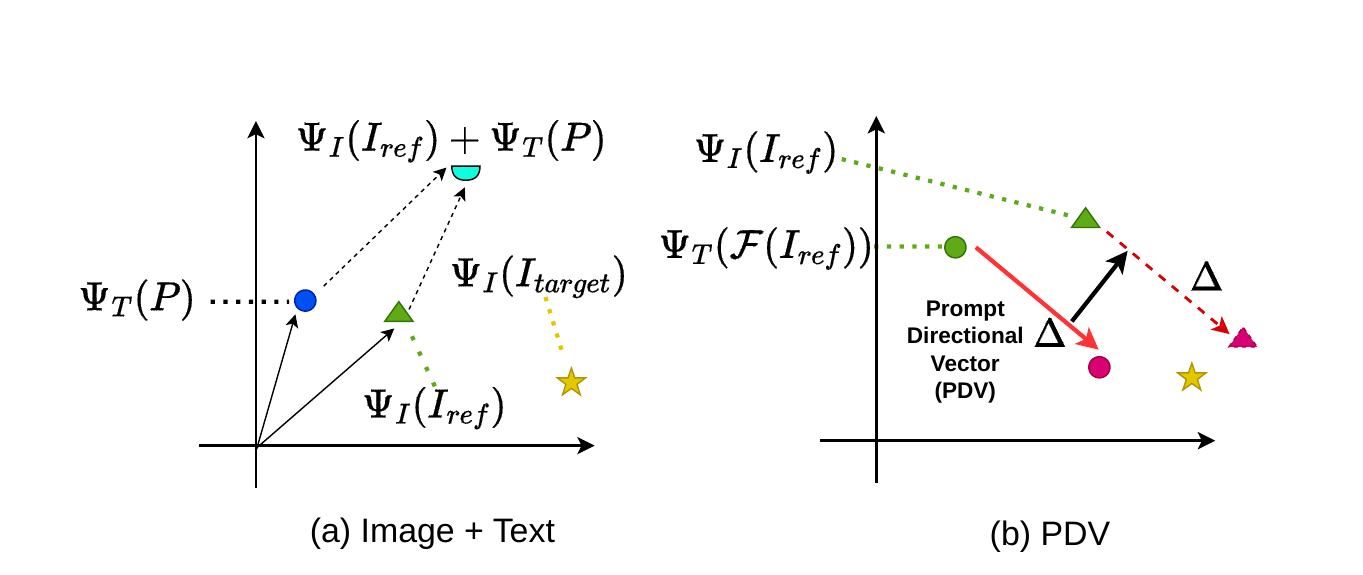}}
	\caption{Comparison of Image + Text (a) vs PDV-I (b).}
	\label{fig:vtvspv}
\end{figure}

\textbf{(2) Prompt Directional Vector for Image (PDV-I)}, which extends the modification principle to visual embeddings. While previous approaches primarily relied on composed text embeddings, experimental results show that direct fusion of image and text features yields inferior performance compared to composed features. We hypothesize that this performance gap arises because the direct text embedding, $\Phi_{T}(P)$, differs significantly from the prompt vector $\Delta_{PDV}$, as illustrated in Figure \ref{fig:vtvspv}. This difference occurs because the semantic meaning of natural language is context-sensitive, where in our case the context is provided by the reference image embedding $\Psi_T(\mathcal{F}(I_{ref}))$. To address this limitation, we propose combining $\Delta_{PDV}$ with visual embeddings. Specifically, we compute the composed visual embedding $\Phi_{PDV-I}$
as $\Psi_{I}(I_{ref}) + \alpha_I\Delta_{PDV}$, where $\Psi_{I}(I_{ref})$ represents the original visual embedding obtained from the reference image, and the same prompt vector obtained via Eq. \ref{eq:pdv} is used to modify this visual representation.

\textbf{(3) Prompt Directional Vector Fusion (PDV-F)}, which calculates the final similarity score between a query and target image which combines both composed embeddings. This fusion embedding, $\Phi_{PDV-F}$, can be defined as,
\begin{equation}
	\Phi_{PDV-F} = (1-\beta)\Phi_{PDV-I} +\beta\Phi_{PDV-T},
\end{equation}
where $\beta$ is a weighting parameter balancing the contribution of the composed visual and textual embeddings.

\subsection{Parameter Tuning and Efficient Searching}
PDV introduces three hyperparameters: $\alpha_T$, $\alpha_I$, and $\beta$. Among these, $\alpha_T$ is most critical, and adjusting it alone typically suffices. Fine-tuning $\alpha_I$ and $\beta$ can further enhance retrieval performance and controllability. Detailed tuning guidelines are in Section \red{S1.1}\footnote{S represent supplementary material}, and our automatic tuning method for $\alpha_I$ is described in Section \red{S1.1.1}.

PDV mainly reduces feature extraction costs in iterative search. We also find that filtering gallery images by prior ranking results lowers ranking costs, another key factor in CIR retrieval. Section \red{S1.2} provides discussion and supporting experiments.

\section{Experiments} 
\label{sec:exp}
\noindent\textbf{Implementation Details.} We utilize the official implementations of four ZS-CIR baseline methods: CIReVL\footnote{https://github.com/ExplainableML/Vision\_by\_Language} and LDRE\footnote{https://github.com/yzy-bupt/LDRE} as representative caption-based feature extraction approaches and Pic2Word\footnote{https://github.com/google-research/composed\_image\_retrieval} and SEARLE\footnote{https://github.com/miccunifi/SEARLE} as representative pseudo tokenization-based methods. All feature extraction processes follow the original implementations provided by baseline methods. However, to calculate $\Delta_{PDV}$, we need text embeddings without prompts, which are not provided in the original implementations. For CIReVL and LDRE, we obtain these embeddings by passing the generated image captions directly to CLIP. For Pic2Word and SEARL, we construct the base text embedding by passing the phrase ``a photo of $\langle$token$\rangle$" to CLIP, where $\langle$token$\rangle$ represents the extracted image token obtained via text inversion.

\noindent\textbf{Datasets and Base Vision-Language Models.} Following previous work, we evaluated our method on a suite of datasets including Fashion-IQ \cite{wu2021fashion}, CIRR \cite{liu2021image} and CIRCO \cite{baldrati2023zero}. Our proposed method is a plug-and-play approach requiring no additional training, leveraging only pre-trained models. For feature extraction, we use three CLIP variants: ViT-B/32, ViT-L/14, and ViT-G/14, and used the same pre-trained weights as used by the baseline methods. For image tokenization, we employ the pre-trained Pic2Word model. 

\subsection{Effect of Using the PDV}
We now explore the impact of the three proposed uses of the PDV: Using the PDV to augment text queries (PDV-T, see Sec. \ref{sec:exp1}), using the PDV to augment image queries (PDV-I, see Sec. \ref{sec:exp2}), and using the PDV in queries that fuse image and text data (PDV-F, see Sec. \ref{sec:exp3}).

\begin{table*}
	\footnotesize
	\centering
	\begin{tabular}{l|l|ccc|cccccccc}
		\hline
		\multicolumn{5}{c|}{\textbf{Fashion-IQ}}  & \multicolumn{2}{c}{\textbf{Shirt}} & \multicolumn{2}{c}{\textbf{Dress}} & \multicolumn{2}{c}{\textbf{Toptee}} & \multicolumn{2}{c}{\textbf{Average}} \\ \hline
		\multicolumn{1}{c|}{Backbone} & \multicolumn{1}{c|}{Method} & $\beta$ & $\alpha_{I}$& $\alpha_{T}$ & R@10 & R@50 & R@10 & R@50 & R@10 & R@50 & R@10 & R@50 \\
		\hline
		\multirow{6}{*}{ViT-B/32} %
		& SEARLE & - & - & - & 24.14 & 41.81 & 18.39 & 38.08 & 25.91 & 47.02 & 22.81 & 42.30 \\
		& SEARLE + \textbf{PDV-F} & 0.9 & 1.1 & 0.9 & \hli{24.83} & 41.71 & \hli{20.13} & \hli{41.40} & \hli{25.96} & \hli{47.17}  & \hli{23.64} & \hli{43.43} \\
		& CIReVL \textdagger &- & -& -& 28.36 & 47.84 & 25.29 & 46.36 & 31.21 & 53.85 & 28.29 & 49.35 \\
		& CIReVL + \textbf{PDV-F} & 0.75 & 1.4 & 1.4 & \hlb{32.88} & \hlb{52.80} & \hlb{32.67} & \hlb{54.49} & \hlb{38.91} & \hlb{61.81} & \hlb{34.82} & \hlb{56.37} \\
		& LDRE \textdagger & - & - & - & 27.38 & 46.27 & 19.97 & 41.84 & 27.07 & 48.78 & 24.81 & 45.63 \\
		& SEIZE \textdagger & - & - & - & \underline{29.38} & \underline{47.97} & \underline{25.37} & \underline{46.84} & \underline{32.07} & \underline{54.78} & \underline{28.94} & \underline{49.86} \\
		\hline
		\multirow{8}{*}{ViT-L/14} & Pic2Word & & & & 25.96 & 43.52 & 19.63 & 40.90 & 27.28 & 47.83 & 24.29 & 44.08 \\
		& Pic2Word + \textbf{PV-F} & 0.8 & 1.0 & 1.0 & \hli{28.21} & \hli{44.55} & \hli{20.92} & \hli{42.24} & \hli{29.02} & \hli{48.90}& \hli{26.05} & \hli{45.23}\\
		& SEARLE & - & - & - & 26.84 & 45.19 & 20.08 & 42.19 & 28.40 & 49.62 & 25.11 & 45.67 \\
		& SEARLE +\textbf{PDV-F} & 0.8 & 1.2 & 1.0 & \hli{28.66} & \hli{46.76} & \hli{23.60} & \hli{46.41} & \hli{31.00} & \hli{52.32} & \hli{27.75} & \hli{48.50} \\
		& CIReVL \textdagger & & & & 29.49 & 47.40 & 24.79 & 44.76 & 31.36 & 53.65 & 28.55 & 48.57 \\
		
		& CIReVL + \textbf{PDV-F} & 0.55 & 1 & 1.3 & \hlb{37.78} & \hlb{54.22} & \hlb{33.61} & \hlb{56.07} & \hlb{41.61} & \hlb{62.16} & \hlb{37.67} & \hlb{57.48} \\
		& LinCIR & - & - & - & 29.10 & 46.81 & 20.92 & 42.44 & 28.81 & 50.18 & 26.82 & 46.49 \\
        & SEIZE & -& -& -& \underline{33.04} & \underline{53.22} & \underline{30.93} & \underline{50.76} & \underline{35.57} & \underline{58.64} & \underline{33.18} & \underline{54.21} \\
		\hline
        \multirow{6}{*}{ViT-G/14} & Pic2Word  & - & - & - & 33.17 & 50.39 & 25.43 & 47.65 & 35.24 & 57.62 & 31.28 & 51.89\\
         & SEARLE  & - & - & - & 36.46 & 55.35 & 28.16 & 50.32 & 39.83 & 61.45 & 34.81 & 55.71\\
		  & CIReVL \textdagger & -& -& -& 33.71 & 51.42 & 27.07 & 49.53 & 35.80 & 56.14 & 32.19 & 52.36 \\
		& CIReVL + \textbf{PV-F} & 0.6 & 1.4 & 1.4 & \hli{41.90} & \hli{58.19} & \hlb{40.70} & \hlb{62.82} & \underline{\hli{48.09}}& \hli{67.77}& \underline{\hli{43.56}}& \hli{62.93}\\
        & LinCIR & - & - & - & \textbf{46.76} & \underline{65.11} & 38.08& 60.88& \textbf{50.48}& \underline{71.09}& \textbf{45.11} & \underline{65.69}\\
        & SEIZE & - & - & - & \underline{43.60} & \textbf{65.42}& \underline{39.61} & \underline{61.02} & 45.94& \textbf{71.12}& 43.05& \textbf{65.85}\\
		\hline
	\end{tabular}
	\caption{Average recall for different methods on Fashion-IQ validation dataset. Yellow highlighting shows PDV improvements over baseline. \colorbox{orange!30}{Orange} = PDV improvements, \textbf{bold} = best, \underline{underline} = second. \textdagger Numbers from original paper.}
	\label{tab:fashion_iq_results}
\end{table*}

\begin{table*}
	\centering
	\footnotesize
	\setlength{\tabcolsep}{4pt}
	\begin{tabular}{l|l|ccc|cccc|cccc|ccc}
		\hline
		\multicolumn{5}{c|}{\textbf{Dataset}} &  \multicolumn{4}{c|}{\textbf{CIRCO}} & \multicolumn{7}{c}{\textbf{CIRR}} \\
		\hline
		\multicolumn{5}{c|}{Metric} & \multicolumn{4}{c|}{mAP@k} & \multicolumn{4}{c|}{Recall@k} &\multicolumn{3}{c}{$R_s$@k} \\
		\hline
		\multicolumn{1}{c|}{Arch} & \multicolumn{1}{c|}{Method} & $\beta$ & $\alpha_I$ & $\alpha_T$ & k=5 & k=10 & k=25 & k=50 & k=1 & k=5 & k=10 & k=50 & k=1 & k=2 & k=3 \\
		\hline
		\multirow{8}{*}{ViT-B/32} 
		& PALAVRA\cite{cohen2022my} \textdagger & -& -& -& 4.61 & 5.32 & 6.33 & 6.80 & 16.62 & 43.49 & 58.51 & 83.95 & 41.61 & 65.30 & 80.94 \\
		& SEARLE \textdagger & -& -&- & 9.35 & 9.94 & 11.13 & 11.84 & 24.00 & 53.42 & 66.82 
		& 89.78 & 54.89 & 76.60 & 88.19 \\
		& SEARLE + \textbf{PDV-F} & 0.9 & 1.4 & 1.2 & \hli{9.99} & \hli{10.50}  & \hli{11.70} & \hli{12.40} & \hli{24.53} & \hli{53.71} & \hli{67.33} & \hli{89.81} & \hli{56.94} & \hli{78.05} & \hli{88.99} \\
		&CIReVL \textdagger & - & - & -& 14.94 & 15.42 & 17.00 & 17.82 & 23.94 & 52.51 & 66.00 & 86.95 & 60.17 & 80.05 & 90.19 \\
		& CIReVL + \textbf{PDV-F} & 0.75 & 1.4 & 1.2 & \hlb{19.90} & \hlb{20.61} & \hlb{22.64} & \hlb{23.52} & \hlb{33.25} & \hlb{64.15} & \hlb{75.23} & \hlb{92.43} & \hlb{65.81} &\underline{\hli{83.76}} &\underline{\hli{92.10}} \\
		& LDRE & -& -& -& 17.81 & 18.04 & 19.73 & 20.67 & 25.69 & 55.52 & 68.77 & 89.86 & 60.10 & 80.58 & 91.04 \\
		& LDRE + \textbf{PDV-F} & 0.75 & 1.4 & 1.4 & 17.80 & \hli{18.78} & \hli{20.61} & \hli{21.56} & \underline{\hli{29.30}} & \underline{\hli{60.39}} & \underline{\hli{72.51}} & \underline{\hli{91.42}} & \hli{63.06} & \hli{82.36} & \hli{91.54} \\
        & SEIZE & -&- &- & \underline{19.04} & \underline{19.64} & \underline{21.55}& \underline{22.49}& 27.47 & 57.42& 70.17 & - & \underline{65.59} & \textbf{84.48}& \textbf{92.77} \\
 		\hline
		\multirow{10}{*}{ViT-L/14}
		& Pic2Word & -& -& -& 6.81 & 7.49 & 8.51 & 9.07 & 23.69 & 51.32 & 63.66 & 86.21 & 53.61 & 74.34 & 87.28 \\
		& Pic2Word + \textbf{PDV-F} & 0.85 & 1.2 & 1.0 & \hli{7.74} &  \hli{8.67} & \hli{9.77} & \hli{10.37} & \hli{23.90} & \hli{51.95} & \hli{64.63} & \hli{87.04} & 53.16  & 74.07 & 87.08\\
		& SEARLE \textdagger & - & - & - & 11.68 & 12.73 & 14.33 & 15.12 & 24.24 & 52.48 & 66.29 & 88.84 & 53.76 & 75.01 & 88.19 \\
		& SEARLE + \textbf{PDV-F} & 0.85 & 1.4 & 1.2 & \hli{12.58} & \hli{13.57} & \hli{15.30} & \hli{16.07} & \hli{25.64} & \hli{53.61} & \hli{66.58} & \hli{88.55} & \hli{55.83} & \hli{76.48} & \hli{88.53} \\
		& CIReVL \textdagger & -& -& -& 18.57 & 19.01 & 20.89 & 21.80 & 24.55 & 52.31 & 64.92 & 86.34 & 59.54 & 79.88 & 89.69 \\
		& CIReVL + \textbf{PDV-F} & 0.75 & 1.4 & 1.2 & \hlb{25.67} & \hlb{26.61} & \underline{\hli{28.81}} & \hlb{29.95} & \hlb{36.24} & \hlb{66.17} & \hlb{76.96} & \hlb{92.29} & \hlb{68.07} & \hlb{85.35} & \hlb{93.47} \\
		& LDRE & -& -& -& 22.32 & 23.75 & 25.97 & 27.03 & 26.68 &55.45  & 67.49 & 88.65 & 60.39 & 80.53 & 90.15 \\
		& LDRE + \textbf{PDV-F} & 0.75 & 1.4 & 1.4 & \hli{25.23} & \hli{26.52} & \hlb{28.94} & \hlb{29.95} & \underline{\hli{30.16}} & \underline{\hli{59.98}} & \underline{\hli{71.90}} & \underline{\hli{90.87}} & \hli{63.66} & \hli{82.87} & \hli{91.57} \\

        & LinCIR & - & - & - &12.59 &13.58 &15.00 &15.85 &25.04 &53.25 &66.68 & - &57.11 &77.37 &88.89\\
        & SEIZE & -& -& -& 24.98 & 25.82 &28.24 &\underline{29.35}& 28.65 &57.16& 69.23& - &\underline{66.22} &\underline{84.05} &\underline{92.34} \\

		\hline
		\multirow{7}{*}{ViT-G/14} & CIReVL \textdagger & -& -& -& 26.77 & 27.59 & 29.96 & 31.03 & 34.65 & 64.29 & 75.06 & 91.66 & 67.95 & 84.87 & 93.21 \\

		& CIReVL + \textbf{PDV-F} & 0.75 & 1.4 & 1.2 & \hli{30.02} & \hli{31.46} & \hli{34.01} & \hli{35.08} & \hli{38.15} &\hli{67.93} & \hli{77.90} & \hli{92.77} & \hli{69.37} & \hli{85.37} & \hli{93.45}  \\
		
		& LDRE & -& -& -& \underline{33.30} & \underline{34.32} & \underline{37.17} & \underline{38.27} & 37.40 & 66.96 & 78.17 & 93.66 & 68.84 & 85.64 & 93.90 \\
		& LDRE + \textbf{PDV-F} & 0.75 & 1.4 & 1.4 & \hlb{34.88} & \hlb{36.41} & \hlb{39.12} & \hlb{40.23} & \hlb{42.51} & \hlb{72.22} & \hlb{81.71} & \hlb{94.94} & \underline{\hli{72.39}} & \underline{\hli{88.34}} & \underline{\hli{94.80}} \\
        & SEARLE & - & - & - & 13.20 &13.85 &15.32 &16.04 & 34.80 & 64.07 & 75.11 &-&68.72 &84.70 &93.23 \\
        & LinCIR & - & - & - & 19.71 &21.01 &23.13 &24.18 &35.25 &64.72 &76.05 & - &63.35 &82.22 &91.98 \\
        & SEIZE & -& -& -& 32.46 & 33.77 &36.46 &37.55 &\underline{38.87} & \underline{69.42} & \underline{79.42} & -&\textbf{74.15} & \textbf{89.23} & \textbf{95.71} \\
		\hline
	\end{tabular}
	\caption{Performance comparison on CIRCO and CIRR test datasets. As in previous works, for CIRCO, mAP@k is reported, while for CIRR both Recall@k and $R_s$@k metrics are used. \colorbox{orange!30}{Orange} = PDV improvements, \textbf{bold} = best, \underline{underline} = second. \textdagger Numbers from original paper.}
	\label{tab:circo_cirr_results}
\end{table*}

\noindent{\textbf{Analysing the PDV for Text (PDV-T)}}
\label{sec:exp1}
To investigate how scaling the prompt vector, $\Delta_{PDV}$, affects retrieval performance with composed text embeddings, we conducted experiments using two zero-shot approaches (CIReVL and Pic2Word) with different backbone networks across three datasets. We evaluated the performance by varying the scaling parameter, $\alpha$ (Eq. \ref{eqn:text_embedding}), from -0.5 to 3 by an interval of 0.1.

\begin{figure*}[!tbh]
	\centering
	\begin{subfigure}[b]{\linewidth}
		\centering
		\includegraphics[width=0.495\linewidth]{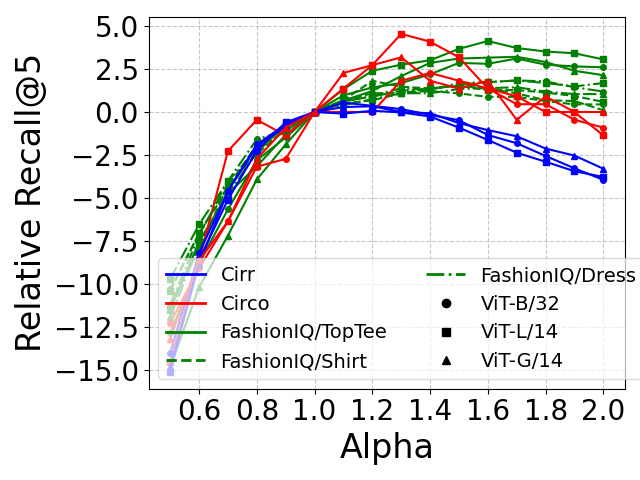}
		\hfil
		\includegraphics[width=0.495\linewidth]{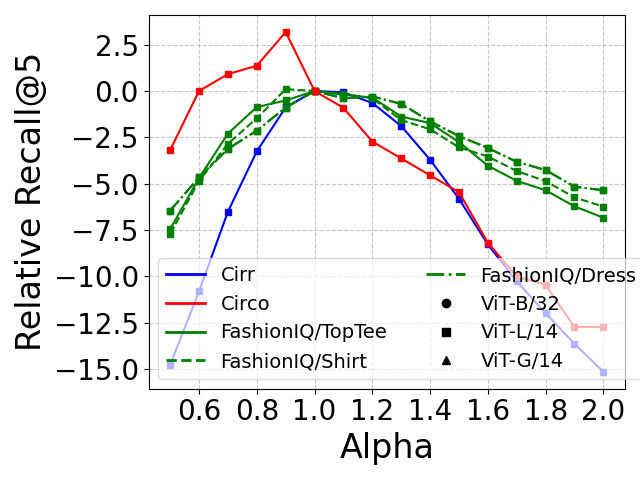}
		\caption{\textbf{PDV-T}: Impact of $\alpha$ scaling on composed text embeddings}
		\label{fig:residual_text_sub}
	\end{subfigure}
	
	\begin{subfigure}[b]{\linewidth}
		\centering
		\includegraphics[width=0.495\linewidth]{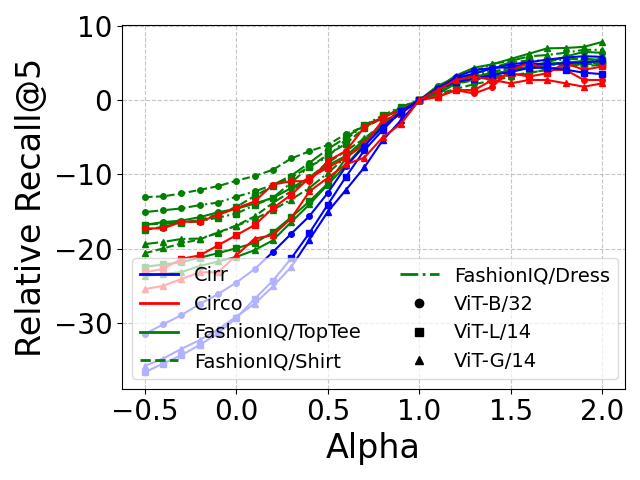}
		\hfil
		\includegraphics[width=0.495\linewidth]{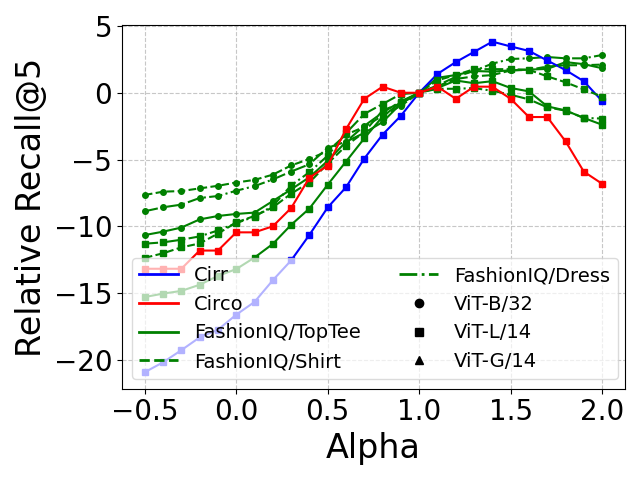}
		\caption{\textbf{PDV-I}: Impact of $\alpha$ scaling on composed image embeddings}
		\label{fig:residual_image_sub}
	\end{subfigure}
	
	\begin{subfigure}[b]{\linewidth}
		\centering
		\includegraphics[width=0.495\linewidth]{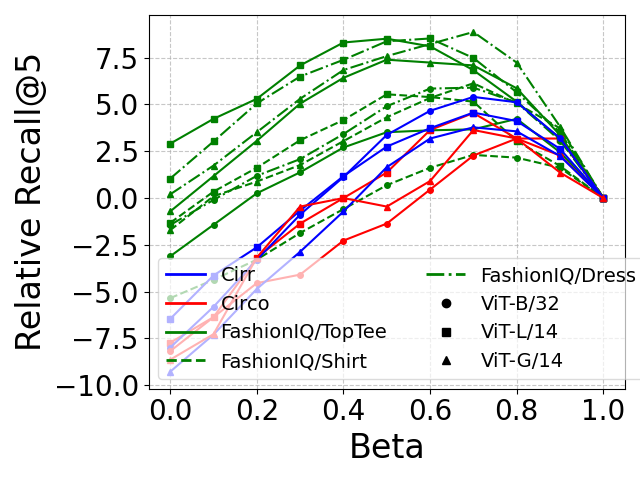}
		\hfil
		\includegraphics[width=0.495\linewidth]{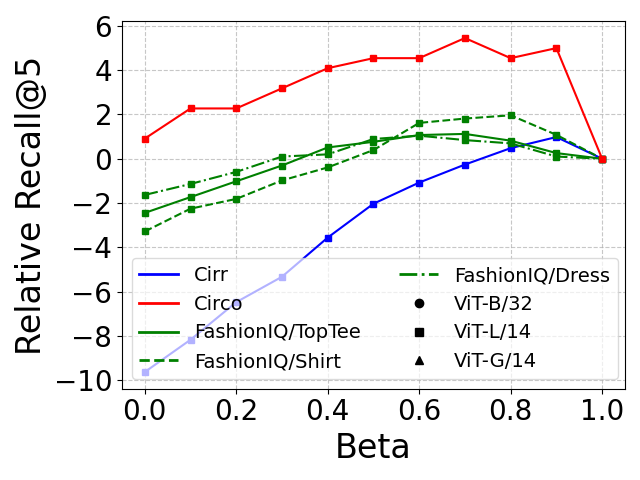}
		\caption{\textbf{PDV-F}: Impact of varying $\beta$ with on composed fused embeddings}
		\label{fig:residual_fusion_sub}
	\end{subfigure}
	\caption{Impact of changing $\alpha$/$\beta$ on Recall@5 performance across different PDV applications. For each row, results are shown for the CIReVL (left) and Pic2Word (right) baseline methods.}
	\label{fig:residual_all}
\end{figure*}

The results are presented in Figure \ref{fig:residual_text_sub}. To account for scale variations across different experiments, we report relative recall values, where a baseline of zero is established at $\alpha=1$. As shown in Figure \ref{fig:residual_text_sub}, varying $\alpha$ leads to significant changes in relative recall performance\footnote{See supplementary material for Recall@10 and Recall@50 figures}. Our analysis reveals method-specific patterns across datasets. With CIReVL, increasing $\alpha$ improves relative recall on both FashionIQ and CIRCO datasets. In contrast, Pic2Word shows no significant improvement on FashionIQ and CIRR when varying $\alpha$, while CIRCO's performance improves when $\alpha$ is reduced to 0.8-1.0. This divergent behavior is fundamentally linked to each method's ability to generate an accurate $\Delta_{PDV}$. As demonstrated in Tables \ref{tab:fashion_iq_results} and \ref{tab:circo_cirr_results}, CIReVL consistently outperforms Pic2Word across various benchmarks, indicating its superior ability to generate a more accuraute composed query, and thus a more accurate $\Delta_{PDV}$. Consequently, increasing $\alpha$ yields greater benefits for CIReVL compared to Pic2Word.

\begin{figure*}[!tbh]
	\centering
	\includegraphics[width=0.95\linewidth]{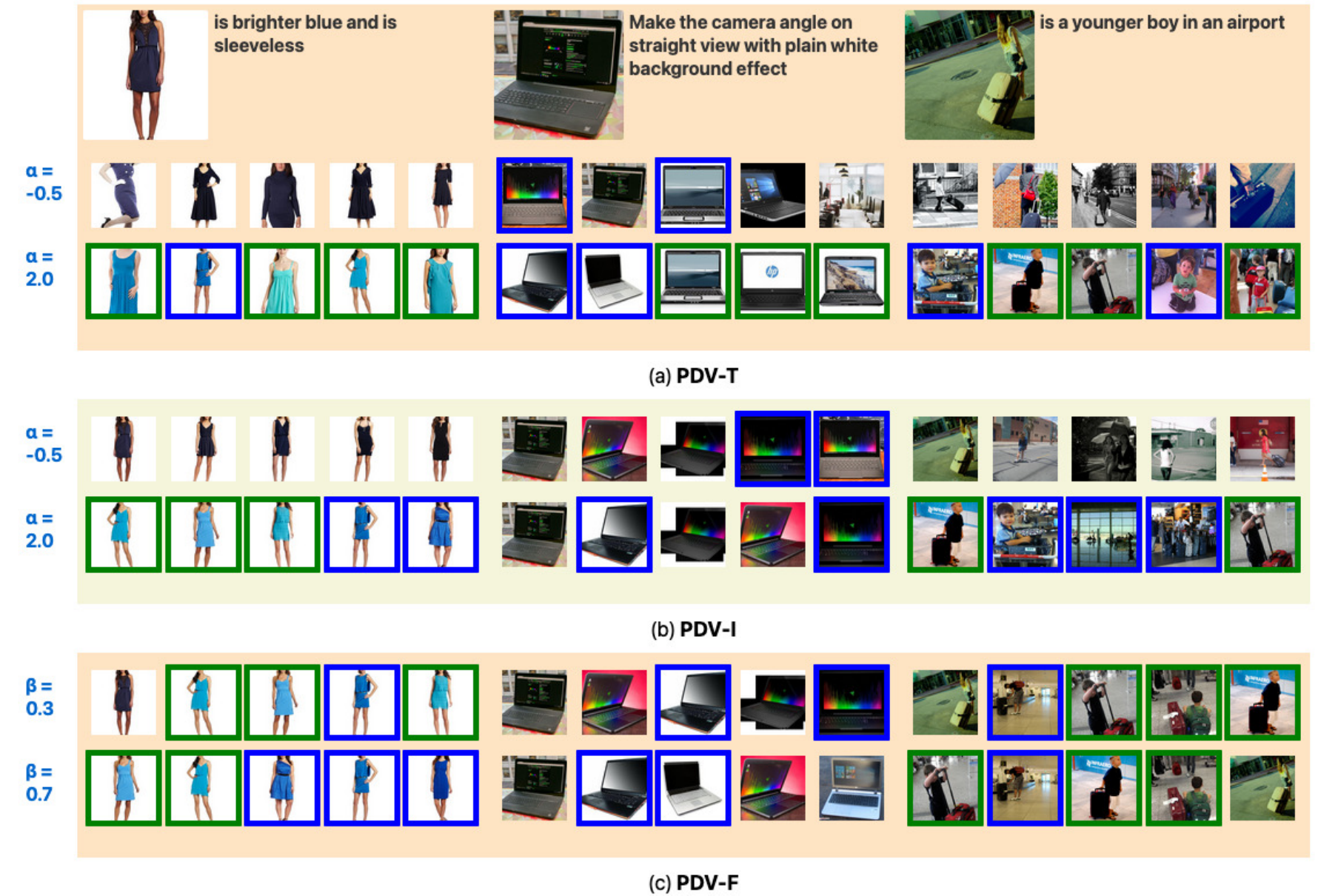}
	\caption{Visualisation of the impact of $\alpha$/$\beta$ scaling on top-5 retrieval results. CIReVL with ViT-B-32 Clip model is the baseline method used. Representative examples with prompts from three datasets: FashionIQ (left), CIRR (middle), and CIRCO (right) are shown at the top. \textbf{\textcolor{boxgreen}{Green}} and \textbf{\textcolor{boxblue}{blue}} bounding boxes indicate true positives and near-true positives, respectively.}
	\label{fig:residual_qual}
	
\end{figure*}

We visualize the top-5 retrieval results using CIReVL with a ViT-B-32 backbone across three datasets (one reference image from each) under varying $\alpha$ values, as shown in Figure \ref{fig:residual_qual}\red{a}. As $\alpha$ increases, the retrieved results show stronger alignment with the prompt. Conversely, when $\alpha$ exceeds 1, the results include semantically related but unseen variations, while $\alpha$ values below 0.5 yields results opposite to the prompt's intent. For instance, ``brighter blue and sleeveless" retrieves ``dark blue with sleeves," ``plain background" yields ``natural/dark background," and ``young boy" returns ``adult" images.

\noindent{\textbf{Analysing the PDV for Image (PDV-I)}}
\label{sec:exp2}
To evaluate whether $\Delta_{PDV}$ enhances the retrieval performance of image embeddings, we conducted experiments following the protocol described in Section~\ref{sec:exp1}. We modified image embeddings by adding $\Delta_{PDV}$ scaled with $\alpha$ values ranging from -0.5 to 2.0, where $\alpha=0$ represents the original image-only embeddings. As shown in Figure \ref{fig:residual_image_sub}, Recall@K exhibits a positive correlation with $\alpha$ for values below 1. This upward trend continues until $\alpha=2.0$ for CIReVL, while Pic2Word's performance peaks when $\alpha$ reaches 1.4.  The performance of PDV-I was evaluated on the FashionIQ, CIRR and CIRCO datasets by comparing it with other visual embedding-based methods, as detailed in Tables \red{S5, S6} in the supplementary material. The results reveal that PDV-I achieved significant improvements over existing approaches.

Following the methodology in Section~\ref{sec:exp1}, we conduct similar visualizations, with results shown in Figure \ref{fig:residual_qual}\red{b}. As with PDV-T, increasing $\alpha$ leads to stronger alignment between retrieved results and the prompt. When $\alpha$ exceeds 0.5, the results exhibit semantic relationships to the query, while $\alpha$ values below 0.5 yield results opposing the prompt's intent.
Notably, PDV-I's top retrievals demonstrate higher visual similarity to reference images compared to PDV-F, as evidenced by the preserved design elements in the clothing item (left) and laptop (middle). This characteristic is particularly valuable for applications include fashion search \cite{wu2021fashion} and logo retrieval \cite{tursun2019component}, where visual similarity plays a crucial role.

\noindent{\textbf{Analysing PDV Fusion (PDV-F)}}
\label{sec:exp3}
Finally, we evaluate the effectiveness of fusing image and text-composed embeddings by varying the fusion parameter, $\beta$, from 0 to 1 while maintaining $\alpha=1$
for both PDV-I and PDV-F. At $\beta=0$, the model relies solely on composed image embeddings, while at $\beta=1$, it uses only composed text embeddings. As shown in Figure \ref{fig:residual_fusion_sub}, the fusion of both embeddings consistently outperforms using either embedding type alone. Optimal retrieval performance is typically achieved when $\beta$ is between 0.4 and 0.8.

We similarly visualize the top-5 retrieved results across different $\beta$ values. As shown in Figure \ref{fig:residual_qual}\red{c}, when $\beta$ is small, the retrieved results maintain high visual similarity to the reference image. Conversely, as $\beta$ exceeds 0.5, the results demonstrate stronger semantic alignment with the prompt.

\subsection{ZS-CIR Benchmark Comparison}

We evaluated PDV-F alongside four baseline approaches (CIReVL, LDRE, Pic2Word, and SEARLE) across three benchmarks. Notably, CIReVL was tested with three different backbones on three datasets, as its models and intermediate results are publicly available. However, for the remaining methods, we conducted evaluations using the models which are publicly available.

Numerical results are presented in Tables \ref{tab:fashion_iq_results} and \ref{tab:circo_cirr_results}. The hyperparameters shown in these tables were selected based on the valid ranges identified in Figure \ref{fig:residual_all}. Since these parameters are intended for user adjustment and users are expected to tune them in response to initial retrieval results, we present the optimal hyperparameters from the three trials.
On the FashionIQ benchmark, PDV-F yields substantial improvements for all baseline approaches, with CIReVL showing particularly strong gains that scale with backbone size. Similarly, all methods demonstrate significant performance improvements on CIRCO and CIRR datasets. Notably, CIReVL achieves larger improvements compared to other methods, with the most substantial gains observed when using small and medium backbone architectures. Our PDV-F implementation within the CIReVL framework consistently outperformed other state-of-the-art methods, including LinCIR and SEIZE, across most evaluation metrics. Similar to SEIZE, PDV-F offers the advantage of being entirely training-free; however, unlike SEIZE, it does not significantly increase feature extraction computational costs. While LinCIR demonstrates exceptional inference speed, it lacks the training-free nature of our approach, requiring dedicated model training before deployment.

\section{Conclusion \& Future Applications}
\label{sec:con}
We introduce the Prompt Directional Vector (PDV), a simple yet effective approach for enhancing Zero-Shot Composed Image Retrieval. PDV captures semantic modifications induced by user prompts without requiring additional training or expensive data collection. Through extensive experiments across multiple benchmarks, we demonstrated three successful applications of PDV: dynamic text embedding synthesis, composed image embedding through semantic transfer, and effective multi-modal fusion. 

Our approach not only improves retrieval performance consistently, but also provides enhanced controllability through the use of scaling factors. PDV serves as a plug-and-play enhancement that can be readily integrated with existing ZS-CIR methods while incurring minimal computational overhead.

We note that PDV's effectiveness correlates strongly with the underlying method's ability to generate accurate compositional embeddings. This insight suggests promising future research directions, including developing more robust compositional embedding techniques and exploring adaptive scaling strategies for PDV. The simplicity and effectiveness of PDV also open possibilities for its application in multi-prompt composed image retrieval (\ie dialogue-based search) and other multi-modal tasks where semantic modifications play a crucial role.

\clearpage

\renewcommand{\appendixname}{Supplement}
\renewcommand{\appendixtocname}{Supplement}
\renewcommand{\appendixpagename}{Supplement}

\def\theequation{S\arabic{equation}}
\renewcommand{\thefigure}{S\arabic{figure}}
\renewcommand{\thesection}{S\arabic{section}}  
\renewcommand{\thetable}{S\arabic{table}}
\setlength\cftsecnumwidth{2.1em}
\setlength\cftsubsecnumwidth{2.8em}

\begin{appendices}

\section{Additional Experiments and Results}

\subsection{Regarding Hyper-parameter Tuning}
In this work, we introduce three parameters: $\alpha_I$, $\alpha_T$, and $\beta$. Once their semantic roles are understood, manually adjusting them becomes intuitive and straightforward: 

\begin{table}[!h]
	\centering
	\begin{tabular}{cp{6.5cm}}
		$\alpha_T\rightarrow$ & Influence of the text prompt on the \underline{semantic} content of the reference image \\
		$\alpha_I\rightarrow$ & Influence of the text prompt on the \underline{visual }content of the reference image \\
		$\beta\rightarrow$ & Controls trade-off between visual and semantic content \\
	\end{tabular}
	\label{tab:parameters}
\end{table}

We also examined the potential for automatic tuning. However, due to several uncertain factors—such as prompt quality, baseline performance on the target dataset, and user expectations—it remains highly challenging to optimise all three parameters automatically. Nevertheless, we find that $\alpha_I$ can be tuned automatically to reduce the gap between PDV-I and PDV-T. In the following subsections, we discuss the manual adjustment of each parameter and the automatic tuning of $\alpha_I$.

\subsubsection{Tuning $\alpha_{T}$: Influence of Text Prompt} 
\begin{figure*}
	\centering
	\includegraphics[width=\linewidth]{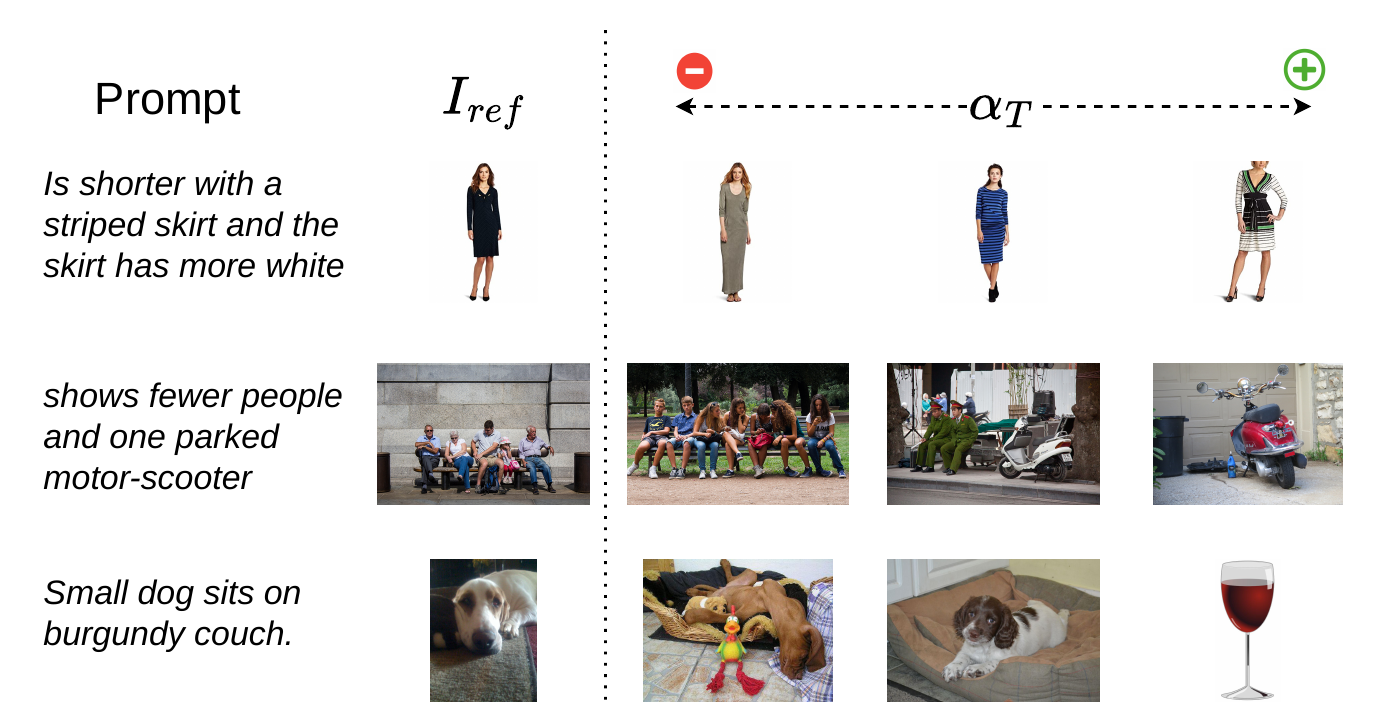}
	\caption{Qualitative results of PDV-T showing the effect of different $\alpha_T$ values. For each query, we display the top-1 retrieval result for three different $\alpha_T$ settings. The middle result uses $\alpha_T = 1$ (baseline), the left result uses a smaller $\alpha_T$ value, and the right result uses a larger $\alpha_T$ value. All $\alpha_T$ values are within the range $[-0.5, 2]$.}
	\label{fig:alpha-t}
\end{figure*}

Among these parameters, the most important is $\alpha_T$, which primarily controls the influence of the text prompt. If a user observes that the semantic changes in the top retrieved results are insufficient, $\alpha_T$ should be increased; conversely, if the changes are too strong, it should be decreased. For example, in the top case of Figure \ref{fig:alpha-t}, when $\alpha_T = 1$, the skirt does not yet display clear white stripes. Increasing $\alpha_T$ produces results with more distinct white stripes. In contrast, in the middle and bottom examples of Figure \ref{fig:alpha-t}, the retrieved results with $\alpha_T = 1$ are already valid. Further increasing $\alpha_T$ in these cases makes the semantic changes too strong, leading the method to return invalid results.

\subsubsection{Manual Tuning $\alpha_{I}$: Influence of Text Prompt}
\begin{figure*}
	\centering
	\includegraphics[width=\linewidth]{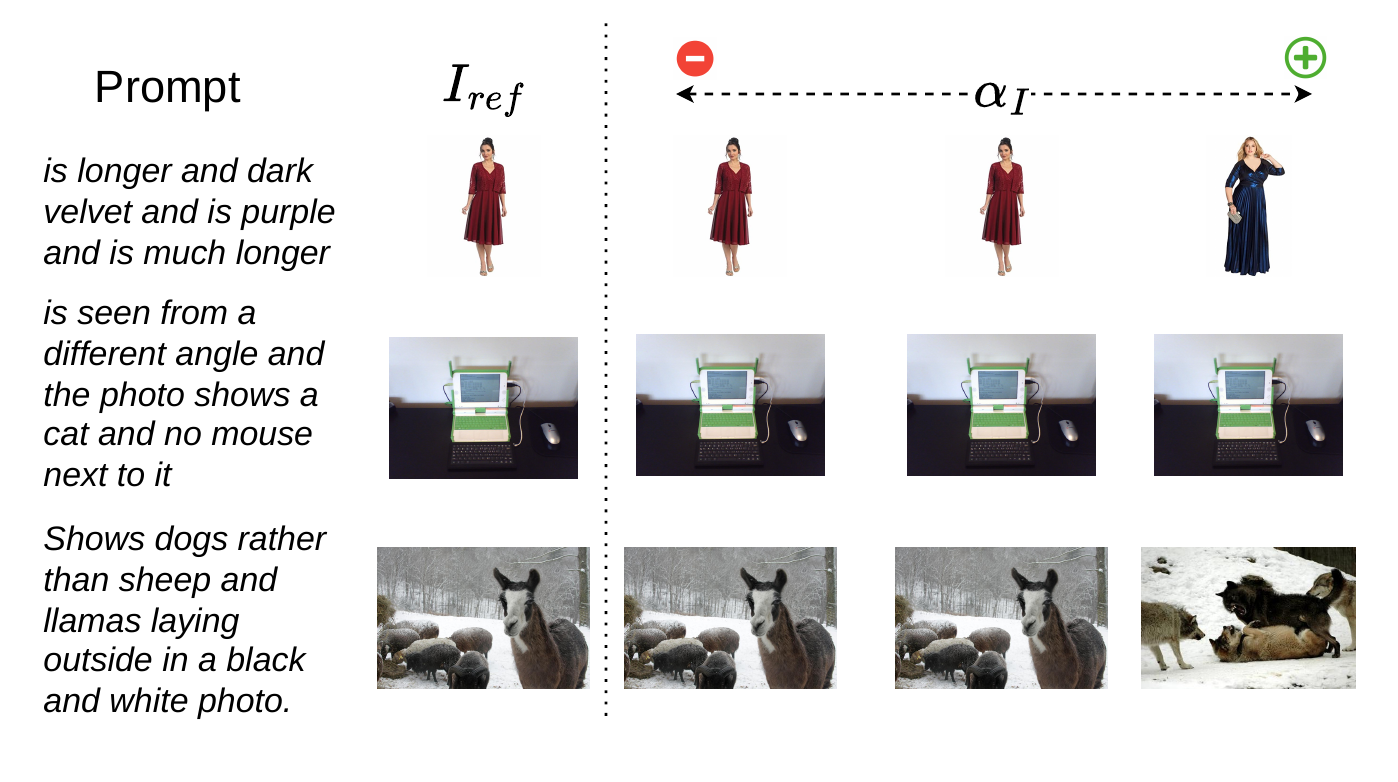}
	\caption{Qualitative results of PDV-I showing the effect of different $\alpha_I$ values. For each query, we display the top-1 retrieval result for three different $\alpha_I$ settings. The middle result uses $\alpha_I = 1$ (baseline), the left result uses a smaller $\alpha_I$ value, and the right result uses a larger $\alpha_I$ value. All $\alpha_I$ values are within the range $[-0.5, 2]$.}
	\label{fig:alpha-i}
\end{figure*}

The role of $\alpha_I$ is similar to that of $\alpha_T$. It also controls the strength of the prompt, but in this case, the composition is with the original visual embedding $\Psi_{I}(I_{ref})$. When $\alpha_I = 0$, PDV-I reduces to content-based image retrieval. From the ablation results shown in Figure \ref{fig:residual_image}, we observe that setting $\alpha_I = 1$ is generally safe, as most methods achieve consistent improvements when $\alpha_I$ is increased from $-0.5$. Nevertheless, further increases in $\alpha_I$ can also be beneficial. Users should continue increasing $\alpha_I$ when the top retrieved results are overly similar to the reference image and fail to incorporate the semantic concepts specified in the prompt. For instance, in the top and bottom examples of Figure \ref{fig:alpha-i}, when $\alpha_I > 1$, the top-1 retrieval results successfully present the semantic elements described in the user prompt.

\subsubsection{Tuning $\beta$: Fusion Factor} 
\begin{figure*}
	\centering
	\includegraphics[width=\linewidth]{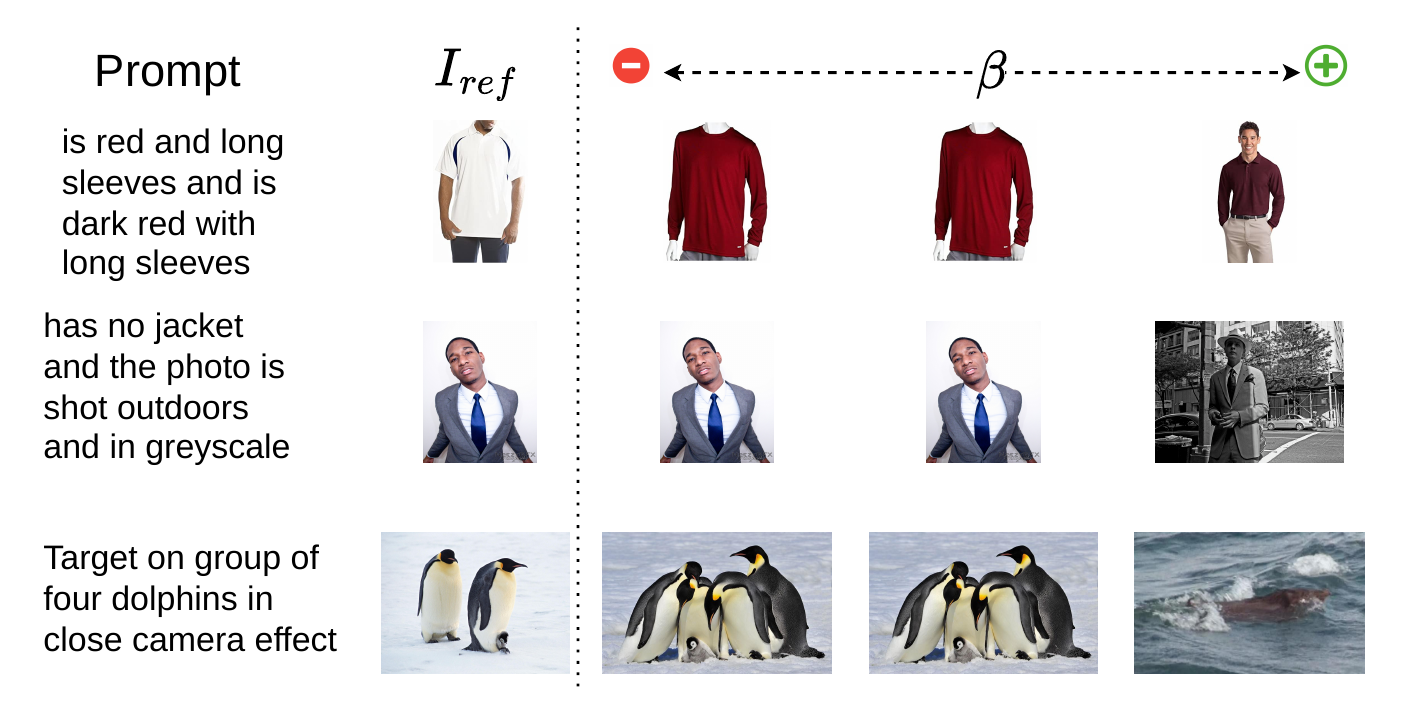}
	\caption{Qualitative results of PDV-F illustrating the effect of different $\beta$ values. For each query, we show the top-1 retrieval result under three settings: $\beta = 0$ (left), $\beta = 0.5$ (middle), and $\beta = 1$ (right).}
	\label{fig:beta}
\end{figure*}

The parameter $\beta$ is used in PDV-F, which fuses PDV-I and PDV-T. Its value ranges from 0 to 1. When $\beta = 1$, PDV-F is equivalent to PDV-T, and when $\beta = 0$, it reduces to PDV-I. From the ablation results shown in Figure \ref{fig:residual_fusion}, we observe that most methods achieve improved performance when $\beta$ lies between 0.6 and 0.9. Beyond performance optimization, $\beta$ also plays a crucial role in balancing retrieval characteristics. As illustrated in Figure \ref{fig:beta}, lower $\beta$ values ($\beta < 0.5$) emphasize visual similarity, producing top results that closely resemble the reference image $I_{ref}$ in terms of appearance. In contrast, higher $\beta$ values ($\beta > 0.5$) prioritize semantic alignment, incorporating conceptual elements described in the text prompt. This provides fine-grained control over whether the retrieval system favors visual fidelity or semantic relevance.

\subsubsection{Automatically Tuning $\alpha_{I}$}
Based on the experimental results, we observe that PDV-T consistently outperforms PDV-I. If the features of PDV-I, denoted as $\Phi_{\text{PDV-I}}$, are more closely aligned with those of PDV-T, $\Phi_{\text{PDV-T}}$, the performance of PDV-I can approach that of PDV-T. Motivated by this observation, we tune the parameter $\alpha_I$ (while keeping $\Phi_{\text{PDV-T}}$ fixed) to minimize the $\ell_2$ distance between $\Phi_{\text{PDV-I}}$ and $\Phi_{\text{PDV-T}}$, as expressed in Equation~\ref{eq:nm}. To determine the optimal value of $\alpha_I$, we employ the Nelder–Mead optimization method \cite{nelder1965simplex}, which is a derivative-free and straightforward approach, making it particularly convenient to implement.

\begin{equation}
	\label{eq:nm}
	\alpha_I = \underset{\alpha}{\arg\min} \; \mathcal{L}\!\left(\Phi_{\text{PDV-T}}, \, \Phi_{\text{PDV-I}}(\alpha)\right),
\end{equation}

To evaluate the effectiveness of the proposed method, we fix $\alpha_T = 1$, making PDV-T equivalent to the baseline, and compare the performance of tuned $\alpha_I$ against the fixed setting $\alpha_I = 1$. We conduct experiments on the FashionIQ dataset using three different methods with multiple backbone architectures. As shown in Table \ref{tab:autoalphaifashioniq}, our approach successfully determines customized $\alpha_I$ values for each setting. In all cases, R@50 shows consistent improvements over the baseline, with the largest gain of 23\% achieved by CIReVL with the ViT-B/32 backbone on the Toptee subset. R@10 also improves steadily in most scenarios, particularly with the CIReVL method. However, for Pic2Word, R@10 decreases by 3.42\% on the Dress subset.

\begin{table*}[h]
	\centering
	\begin{tabular}{l|l|c|cc|cc|cc}
		\hline
		\multirow{2}{*}{\textbf{Backbone}} & \multirow{2}{*}{\textbf{Method}} & \multirow{2}{*}{{$\alpha_{I}$}} 
		& \multicolumn{2}{c|}{\textbf{Shirt}} 
		& \multicolumn{2}{c|}{\textbf{Dress}} 
		& \multicolumn{2}{c}{\textbf{Toptee}} \\
		\cline{4-9}
		& & & R@10 & R@50 & R@10 & R@50 & R@10 & R@50 \\
		\hline
		\multirow{2}{*}{ViT-B/32} 
		& SEARLE   & 1.57/1.65/1.58 & 15.68\% & 15.62\% & 20.04\% & 20.51\% & 14.52\% & 11.18\% \\
		& CIReVL   & 1.92/2.24/2.02 & 25.65\% & 18.82\% & 24.95\% & 16.86\% & 27.81\% & 23.11\% \\
		\hline
		\multirow{3}{*}{ViT-L/14} 
		& CIReVL   & 1.90/2.16/1.95 & 12.96\% & 9.58\%  & 15.12\% & 10.50\% & 17.70\% & 12.80\% \\
		& Pic2Word & 1.47/1.46/1.48 & 0.00\%  & 3.09\%  & -3.42\% & 2.07\%  & 0.00\%  & 0.74\%  \\
		& SEARLE   & 1.67/1.82/1.73 & 5.84\%  & 10.62\% & 16.57\% & 9.41\%  & 6.15\%  & 9.86\%  \\ 
		\hline
		\multirow{1}{*}{ViT-G/14} 
		& CIReVL   & 1.50/1.63/1.53 & 10.43\% & 6.61\%  & 19.15\% & 14.30\% & 17.51\% & 12.14\% \\
		\hline
	\end{tabular}
	\caption{Performance differences with automatic $\alpha_I$ tuning compared to the fixed setting $\alpha_I = 1$ on the FashionIQ datasets. The $\alpha_I$ column reports the tuned values for the Shirt, Dress, and Toptee subsets.}
	\label{tab:autoalphaifashioniq}
\end{table*}

\subsection{Efficient Retrieval with PDV}
PDV is designed to enhance the retrieval performance of baseline methods in a subsequent search, triggered when an initial query fails, without incurring the high computational cost typically associated with iterative search processes.

The computational bottleneck in Zero-Shot Composed Image Retrieval (ZS-CIR) systems stems from two primary operations: feature extraction and similarity ranking.

Regarding feature extraction, the cost is dictated by the model employed. Recent ZS-CIR approaches rely on large vision-language models, whose feature extraction overhead is significant, as detailed in Table \ref{tab:runtime}. In contrast, PDV generates new features for subsequent trials by building upon the embeddings from the initial retrieval. This process involves only efficient scalar multiplications and matrix additions, making its per-trial feature extraction cost nearly negligible. The only substantial computational overhead is the one-time initial calculation of the reference image embedding, $\Psi_{I}(I_{ref})$.

The cost of similarity ranking, on the other hand, is primarily a function of the feature dimension and the gallery size. While the feature dimension is fixed by the base model, the gallery size can be reduced for subsequent searches. To improve efficiency, we integrate a simple filtering strategy with PDV: items whose distance from the query exceeds a predefined threshold are removed from the gallery for subsequent ranking. While not unique to PDV, we believe this is the first discussion of such an optimization in a ZS-CIR context.

We evaluated this approach on the FashionIQ dataset using two baseline methods, CIReVL and Pic2Word. As shown in Table \ref{tab:filter}, for CIReVL, a threshold of 0.8 filters out over 80\% of the gallery items while degrading the R@50 metric by at most 1.31\%. For Pic2Word, a threshold of 0.75 filters out over 68\% of the gallery with a maximum R@50 decrease of 3.73\%. These results demonstrate that PDV, combined with gallery filtering, offers a highly effective trade-off, significantly accelerating retrieval speed while maintaining competitive accuracy.

\begin{table}[!t]
	\centering
	\footnotesize
	\begin{tabular}{l|c|c}
		\hline
		\textbf{Method} & \multicolumn{2}{c}{\textbf{Feature Extraction Time (Sec.)}}\\\hline
		&  Initial   & Retrial \\\hline
		Pic2Word        & 0.02 & 0.02 \\
		\quad + PDV			 & 0.03 &  0.00 \\
		LinCIR          & 0.02 & 0.02 \\
		\quad + PDV           & 0.03 & 0.00 \\
		KEDs            & 0.03 & 0.04 \\
		CIReVL (2 Captions)  & 1.23 & 1.23   \\
		\quad + PDV				& 1.24 & 0.00 \\
		LDRE (20 Captions)  & 17.30  & 17.30    \\
		\quad + PDV			& 17.31  & 0.00    \\\hline
	\end{tabular}
	\caption{Comparison of computation efficiency of the baseline ZS-CIR approaches on NVIDIA A100 GPU. }
	\label{tab:runtime}
\end{table}

\begin{table*}[!th]
	\centering
	\begin{tabular}{l l l c c c c}
		\hline
		\textbf{Backbone} & \textbf{Method} & \textbf{Dataset} & \textbf{Threshold} & \textbf{Filtered Ratio} & \textbf{Change in R@10} & \textbf{Change in R@50} \\
		\hline
		\multirow{3}{*}{ViT-B/32} & \multirow{3}{*}{CIReVL + PDV-F} 
		& Toptee & 0.8  & 90.85\% & -2.88\% & -1.31\% \\
		& & Dress  & 0.8  & 82.31\% & 0.00\%  & -0.73\% \\
		& & Shirt  & 0.8  & 87.96\% & -1.22\% & -1.31\% \\
		\hline
		\multirow{3}{*}{ViT-L/14} & \multirow{3}{*}{Pic2Word + PDV-F} 
		& Toptee & 0.75 & 85.62\% & -1.24\% & -0.94\% \\
		& & Dress  & 0.75 & 68.79\% & 0.00\%  & 0.12\%  \\
		& & Shirt  & 0.75 & 88.36\% & -1.91\% & -3.73\% \\
		\hline
	\end{tabular}
	\caption{Performance changes of PDV-F after applying filtering to the initial retrieval results on the FashionIQ dataset.}
	\label{tab:filter}
\end{table*}

\subsection{$\phi$ Angles of the Baselines}
In Section 3.3 of the main paper, we discussed that PDV's performance is highly correlated with baseline performance and provided theoretical justification through simulation results as shown in Figure \ref{fig:theta}. We have introduced a new parameter $\phi$, which is the angle between the calculated prompt directional vector $\Delta_{\mathrm{PDV}}$ and the ground truth prompt directional vector $\Delta_{\mathrm{GT}}$. When $\phi$ is small, adjusting the parameter $\alpha$ can effectively reduce $\theta$ which is the angle between the target embedding vector $\Psi_{I}(I_{target})$ and the composed embedding vector $\Psi_T(\mathcal{F}(I_{ref}, P))$.

\begin{figure*}[!tbh]
	\centering
	\includegraphics[width=0.8\linewidth]{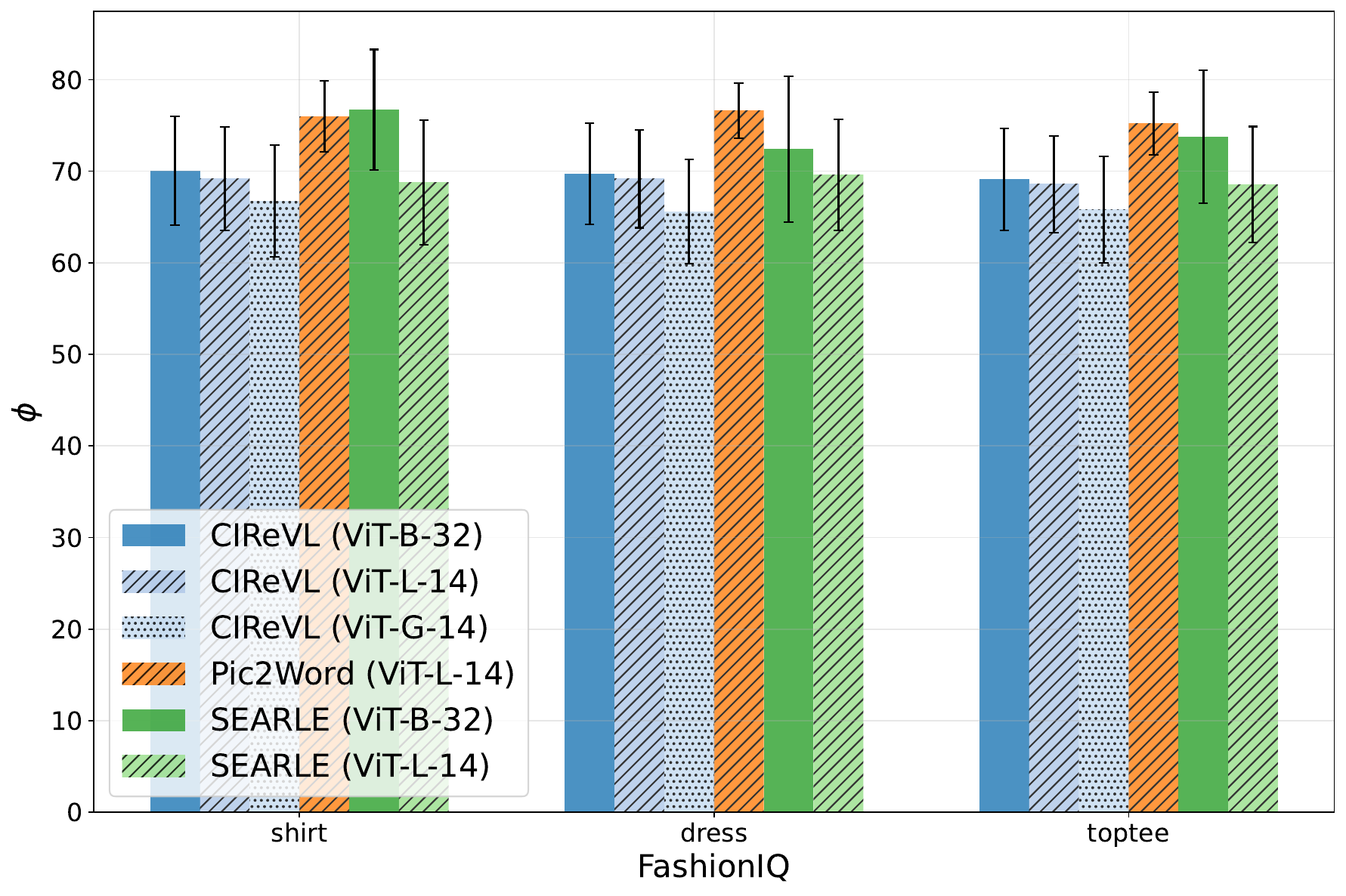}
	\caption{Performance Comparison: $\phi$ Angles by Method and Model Across FashionIQ Datasets}
	\label{fig:phi}
\end{figure*}

Here, we present the actual $\phi$ values for three baseline methods—CIReVL, Pic2Word, and SEARLE—using different backbones across three subdatasets of the FashionIQ dataset. Figure \ref{fig:phi} presents the $\phi$ values for three baseline methods—CIReVL, Pic2Word, and SEARLE—across the FashionIQ subdatasets. While we observed the expected trend of stronger models exhibiting smaller $\phi$ values, we were surprised to find that the state-of-the-art CIReVL baseline maintains a large $\phi$ angle of approximately $65^\circ$. This finding suggests that PDV has not yet been evaluated with optimal baseline models, despite already yielding considerable gains with CIReVL. We therefore anticipate that PDV will be even more effective when paired with future baselines that achieve a smaller $\phi$ (ideally $<60^\circ$), the regime where, according to Figure \ref{fig:theta-b}, our method is most impactful.

\subsection{Additional Quantitative Results}
In this supplementary section, we present additional quantitative results that were omitted from the main paper due to space constraints. 

\subsubsection{Ablation Analysis}
While Figure 3 in the main paper illustrates the effects of scaling factor $\alpha$ and fusion factor $\beta$ on Recall@5 performance across various PDV applications, Figures \ref{fig:residual_text}, \ref{fig:residual_image}, and \ref{fig:residual_fusion} present complementary results for Recall@10 and Recall@50 metrics.

The Recall@10 and Recall@50 results demonstrate consistent trends with the Recall@5 findings presented in the main paper, thus validating our conclusions across multiple evaluation metrics.

\subsubsection{PDV-I Results}
We also provide additional PDV-I results achieved on the validation set of the FashionIQ dataset, as shown in Tables \ref{tab:fashion_iq_results_pdv-i} and \ref{tab:circo_cirr_results_pdv-I}. PDV-I also achieved significant improvements over existing approaches that directly leverage image embeddings for retrieval.

Lastly, we provide a detailed visualization of the impact of $\alpha$/$\beta$ scaling on top-5 retrieval results. Figure \ref{fig:sub_residual_qual} illustrates the performance of CIReVL with the ViT-B-32 CLIP model across three different datasets.

\subsubsection{Zero-shot methods with PDV Vs. Supervised Methods}

We compared state-of-the-art zero-shot composed image retrieval (ZS-CIR) methods enhanced with PDV against supervised methods on the FashionIQ and CIRR datasets. Our evaluation included early supervised methods such as TIRG \cite{vo2019composing}  and ARTEMIS cite{delmas2022artemis}, as well as recent state-of-the-art approaches like CCIN \cite{Tian_2025_CVPR} and SPRC \cite{bai2023sentence}.
The comparative results presented in Table \ref{tab:pdvvssup} demonstrate that PDV achieves remarkably competitive performance against supervised methods. The PDV-based approaches significantly outperform early supervised baselines, with substantial improvements over TIRG (41.90\% vs 14.13\% R@10 on FashionIQ Dress) and ARTEMIS \cite{delmas2022artemis} (41.90\% vs 25.68\%). While PDV methods do not quite match the performance of recent state-of-the-art approaches like CCIN  and SPRC, the performance gap remains relatively modest—typically within 7-9 percentage points on FashionIQ and approximately 8-9 points on CIRR's mean score (72.85\% vs 81.66\% for CCIN).
This narrow performance gap is particularly noteworthy given that ZS-CIR methods with PDV operate without human-annotated training data. These results suggest that unsupervised approaches are reaching a level of effectiveness that positions them as viable alternatives to supervised methods.

\begin{table*}[ht]
	\footnotesize
	\centering
	\begin{tabular}{l|cc|cc|cc|c|c|c|c|c|c}
		\hline
		\multirow{3}{*}{Method} & \multicolumn{7}{c|}{FashionIQ} & \multicolumn{5}{c}{CIRR} \\
		\cline{2-13}
		& \multicolumn{2}{c|}{Dress} & \multicolumn{2}{c|}{Shirt} & \multicolumn{2}{c|}{Toptee} & \multirow{2}{*}{Mean} & \multicolumn{5}{c}{} \\
		& R@10 & R@50 & R@10 & R@50 & R@10 & R@50 & & R@1 & R@5 & R@10 & R@50 & Mean \\
		\hline
		TIRG \cite{vo2019composing}      & 14.13 & 34.61 & 13.10 & 30.91 & 14.79 & 34.37 & 23.66 & 14.61 & 48.37 & 64.08 & 90.03 & 54.27 \\
		ARTEMIS \cite{delmas2022artemis}   & 25.68 & 51.05 & 21.57 & 44.13 & 25.89 & 55.06 & 37.68 & 16.96 & 46.10 & 61.31 & 87.73 & 53.03 \\
		CLIP4CIR \cite{Baldrati_2022_CVPR}   & 33.81 & 59.40 & 39.99 & 60.45 & 41.41 & 65.37 & 50.03 & 38.53 & 69.98 & 81.86 & 95.93 & 71.58 \\
		CompoDiff \cite{gu2023compodiff} & 40.65 & 57.14 & 36.87 & 57.39 & 43.93 & 61.17 & 49.53 & 22.35 & 54.36 & 73.41 & 91.77 & 60.47 \\
		TG-CIR \cite{wen2023target}    & 45.22 & 69.66 & 52.60 & 72.52 & 56.14 & 77.10 & 58.05 & 45.25 & 78.29 & 87.16 & 97.30 & 77.00 \\
		SPRC \cite{bai2023sentence}       & 48.83 & 72.09 & 53.83 & 74.14 & \textbf{58.13} & \textbf{78.58} & 64.27 & 51.96 & 82.12 & 89.74 & 97.69 & 80.37 \\
		CCIN \cite{Tian_2025_CVPR}  & \textbf{49.38} & \textbf{72.58} & \textbf{55.93} & \textbf{74.14} & 57.93 & 77.56 & \textbf{64.59} & \textbf{53.41} & \textbf{84.05} & \textbf{91.17} & \textbf{98.00} & \textbf{81.66} \\
		\hline
		\textbf{CIReVL + PDV}            & 41.90 & 58.19 & 40.70 & 62.82 & 48.09 & 67.77 & 53.25 & 38.15 &67.93 & 77.90 & 92.77 & 69.19\\
		\textbf{LDRE + PDV} & - & - & - & - & - & - & - & 42.51 & 72.22  & 81.71 & 94.94 & 72.85\\
		\hline
	\end{tabular}
	\caption{Comparison of PDV with supervised methods on FashionIQ and CIRR datasets.}
	\label{tab:pdvvssup}
\end{table*}

\cleardoublepage
\section{PDV Algorithm and Code}
\begin{figure*}[!b]
	\centering
	\includegraphics[width=0.8\linewidth]{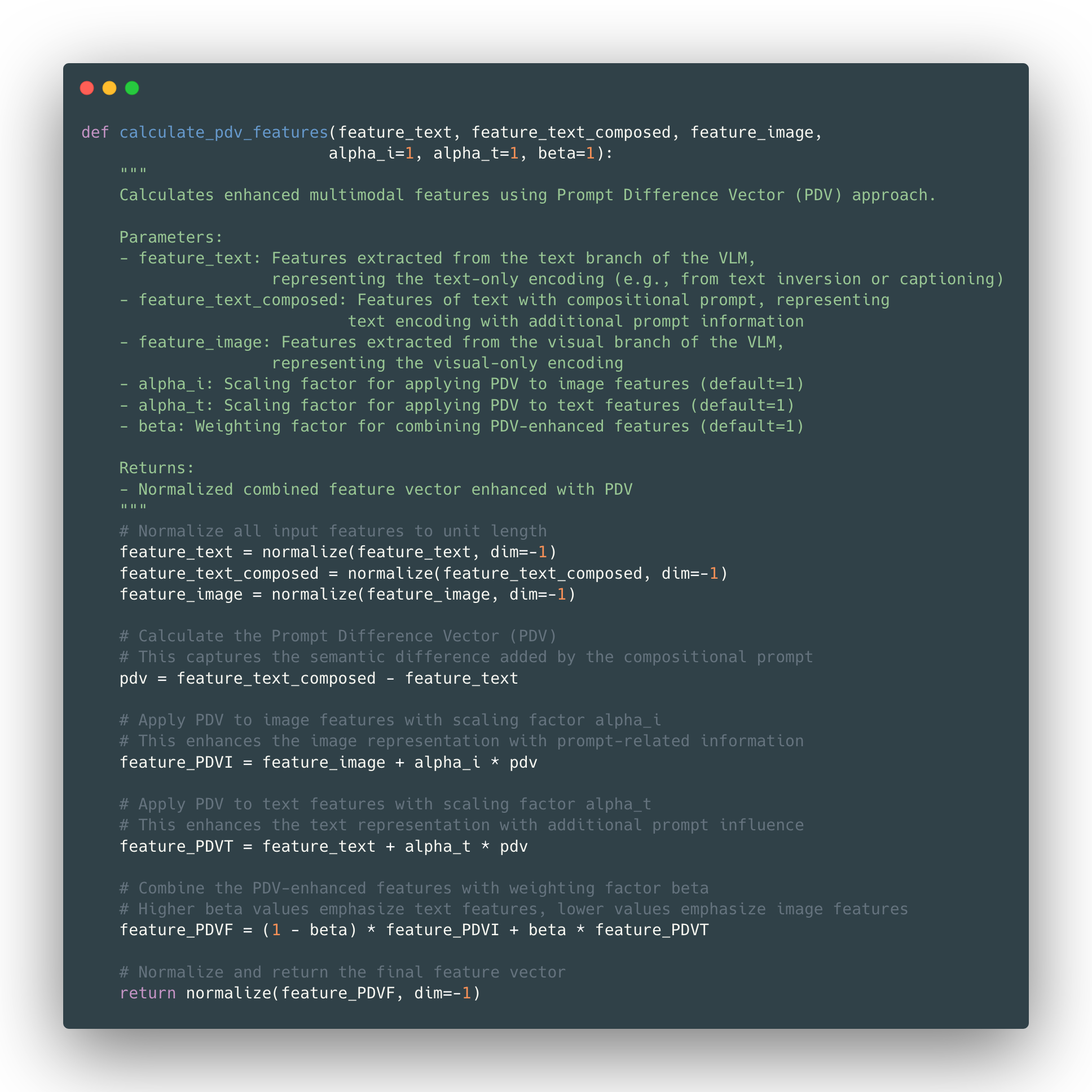}
	\caption{Python function for calculating PDV features.}
	\label{fig:pdv-code}
\end{figure*}

\begin{figure*}
	\centering
	\includegraphics[width=0.9\linewidth]{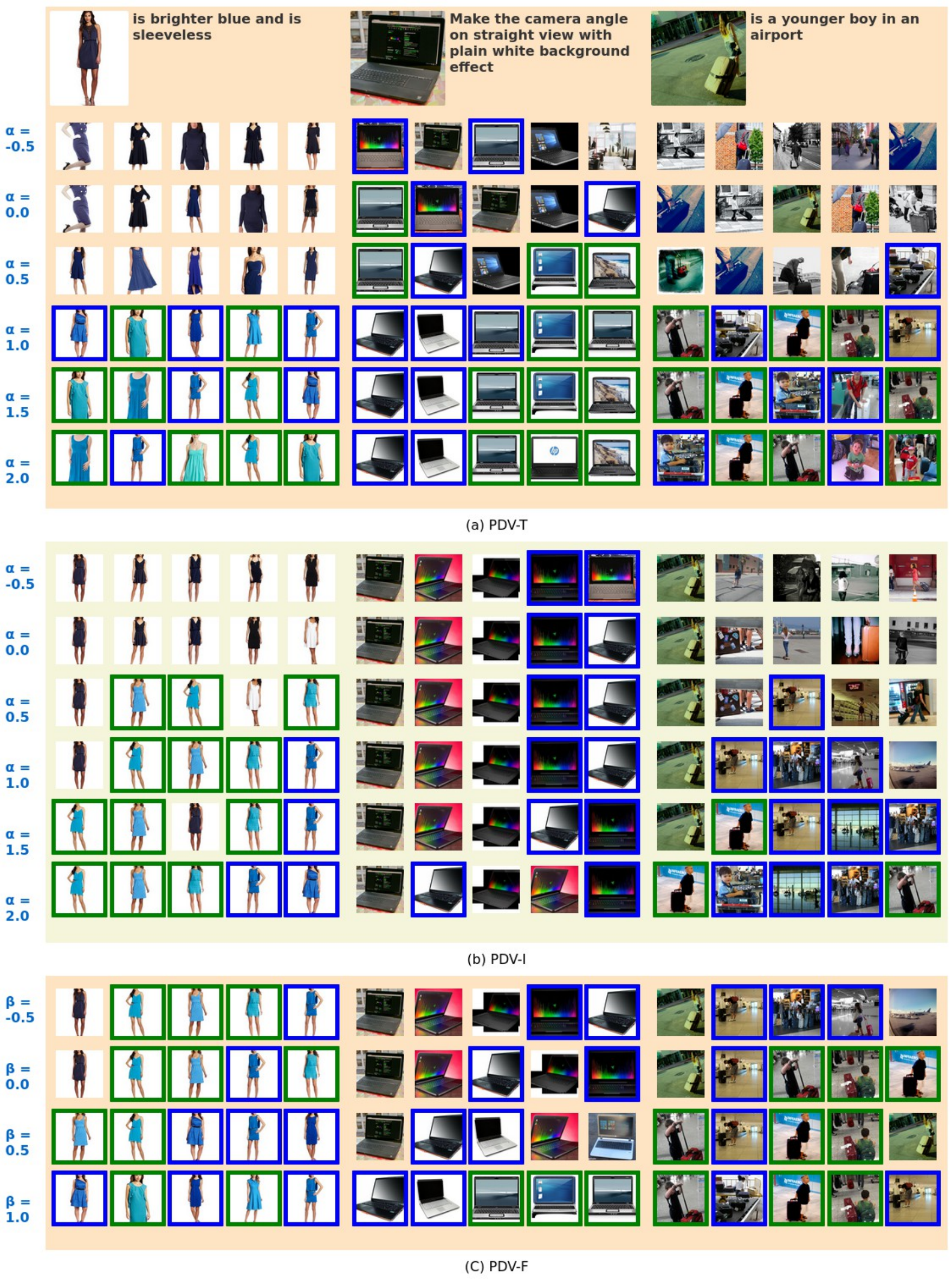}
	\caption{Visualisation of the impact of $\alpha$/$\beta$ scaling on top-5 retrieval results. CIReVL with ViT-B-32 Clip model is the baseline method used. Representative examples with prompts from three datasets: FashionIQ (left), CIRR (middle), and CIRCO (right) are shown at the top. \textbf{\textcolor{boxgreen}{Green}} and \textbf{\textcolor{boxblue}{blue}} bounding boxes indicate true positives and near-true positives, respectively.}
	\label{fig:sub_residual_qual}
	
\end{figure*}

\begin{figure*}
	\centering
	\includegraphics[width=0.45\linewidth]{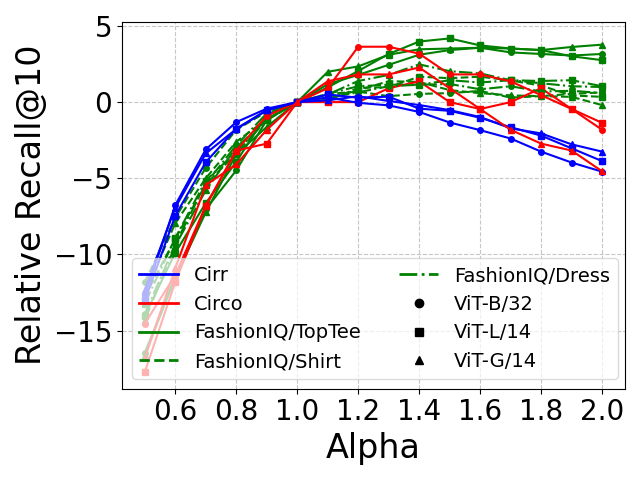}
	\includegraphics[width=0.45\linewidth]{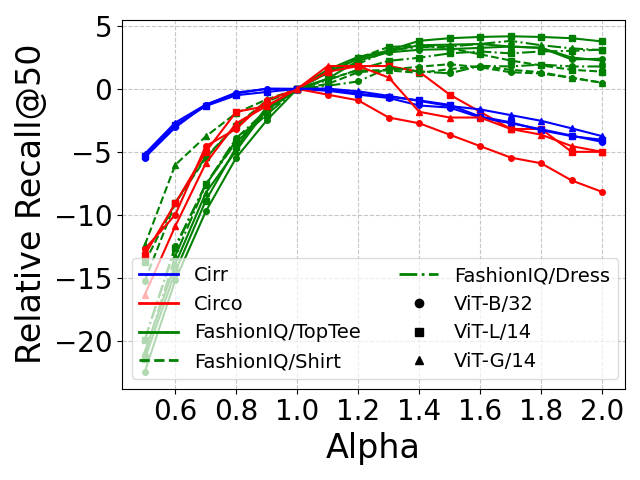}
	\\
	\includegraphics[width=0.45\linewidth]{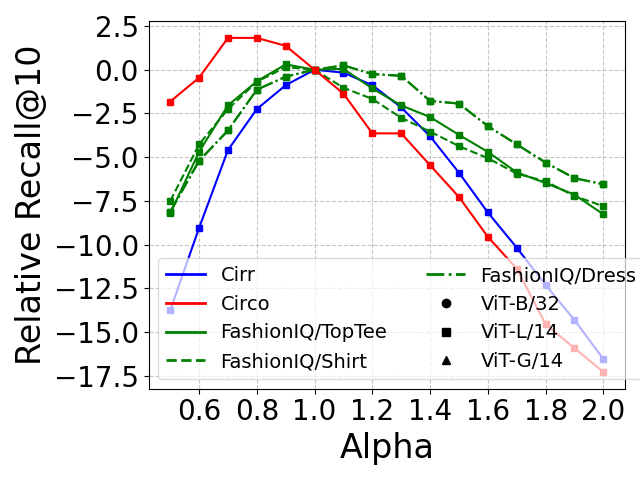}
	\includegraphics[width=0.45\linewidth]{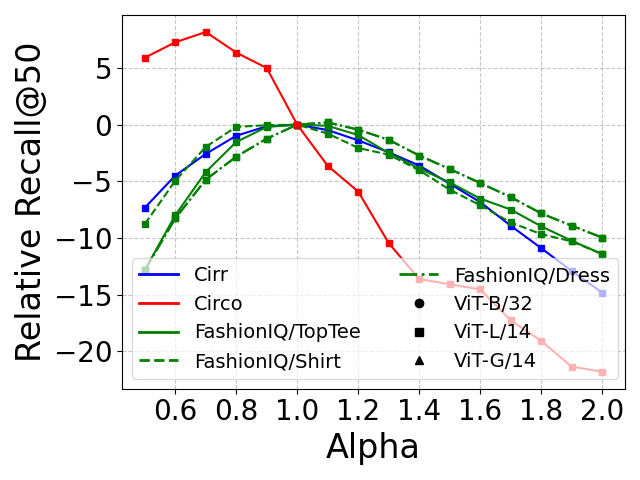}
	\\
	\includegraphics[width=0.45\linewidth]{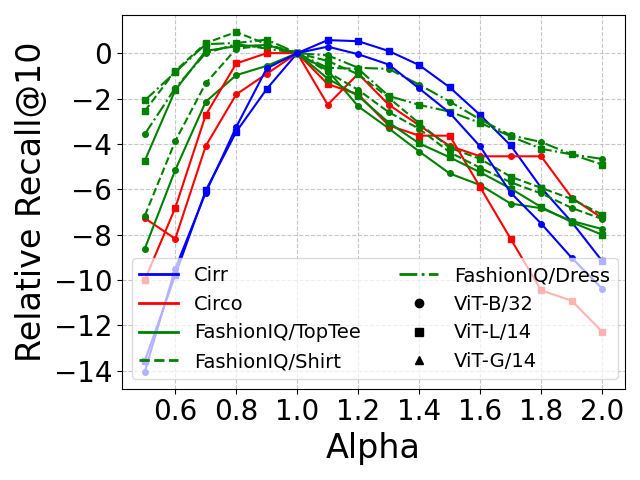}
	\includegraphics[width=0.45\linewidth]{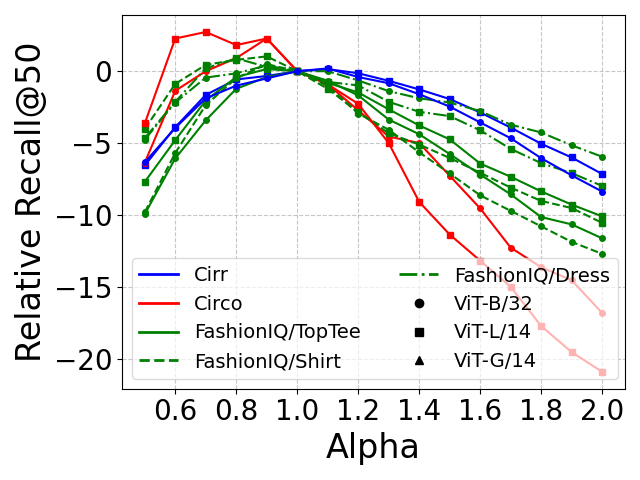}
	\caption{PDV-T: Impact of $\alpha$ scaling on Recall@10 (left) and Recall@50 (right) performance. Results shown for three baseline methods: CIReVL (top), Pic2Word (middle) and SEARLE (bottom).}
	\label{fig:residual_text}
\end{figure*}

\begin{figure*}
	\centering
	\includegraphics[width=0.45\linewidth]{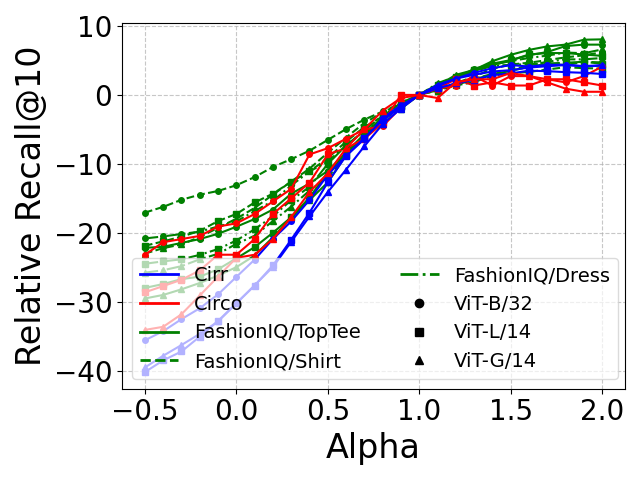}
	\includegraphics[width=0.45\linewidth]{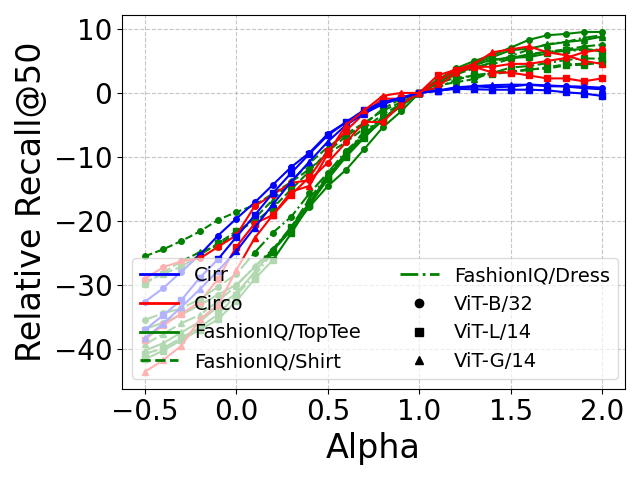}
	\\
	\includegraphics[width=0.45\linewidth]{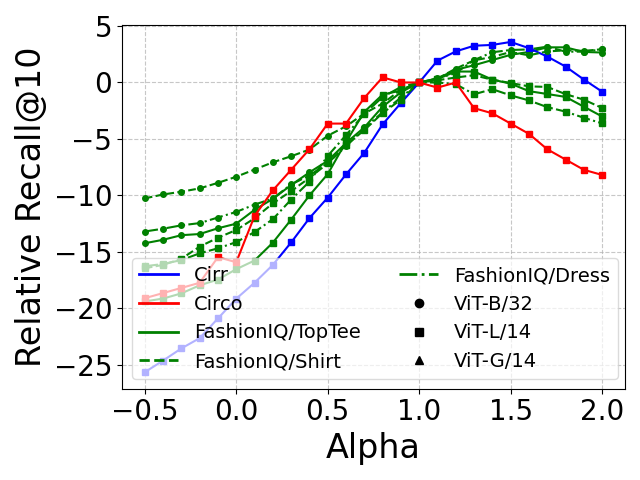}
	\includegraphics[width=0.45\linewidth]{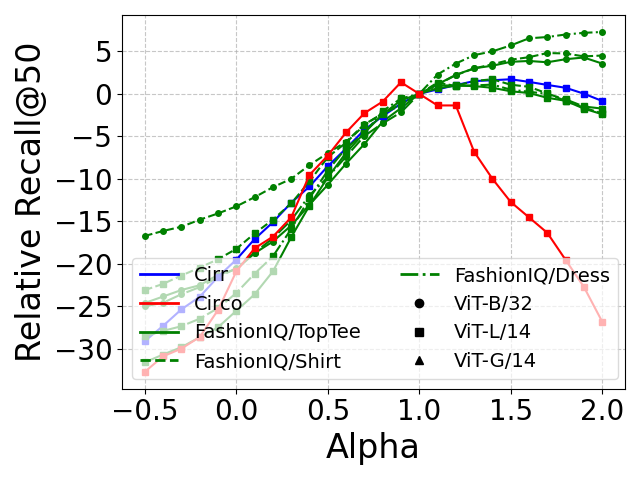}
	\\
	\includegraphics[width=0.45\linewidth]{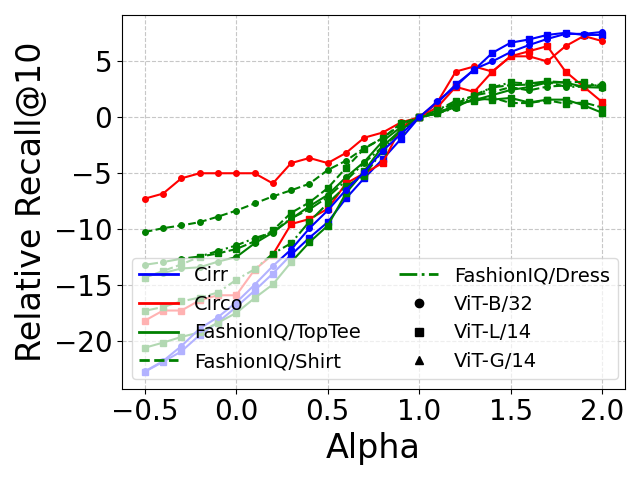}
	\includegraphics[width=0.45\linewidth]{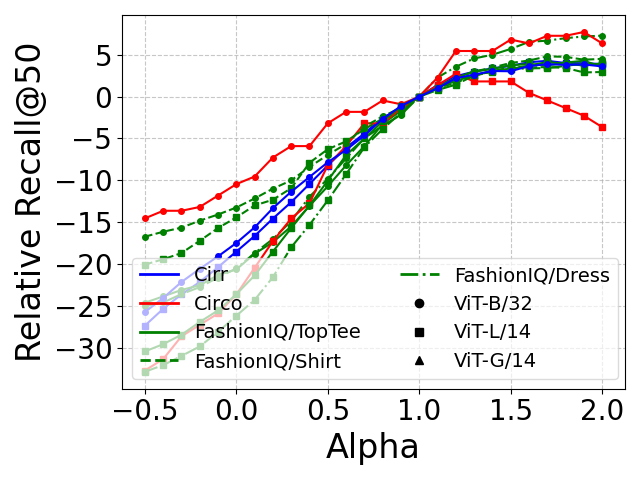}
	\caption{PDV-I: Impact of $\alpha$ scaling on Recall@10 (left) and Recall@50 (right) performance. Results shown for three baseline methods: CIReVL (top), Pic2Word (middle) and SEARLE (bottom).}
	\label{fig:residual_image}
\end{figure*}

\begin{figure*}
	\centering
	\includegraphics[width=0.45\linewidth]{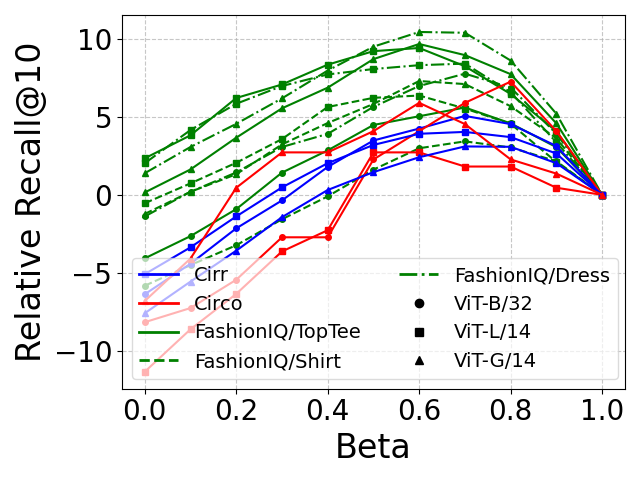}
	\includegraphics[width=0.45\linewidth]{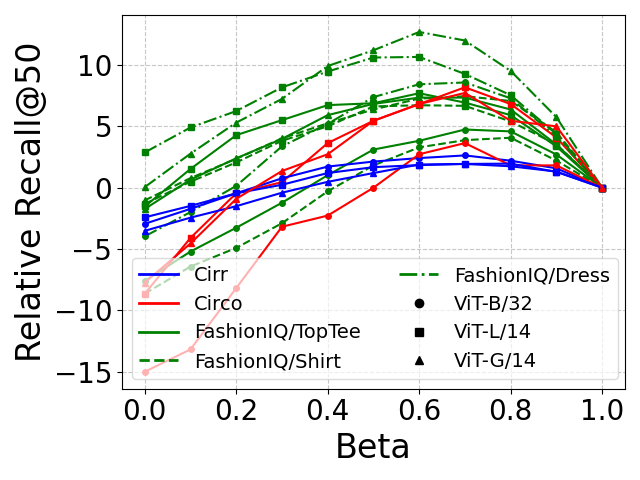}
	\\
	\includegraphics[width=0.45\linewidth]{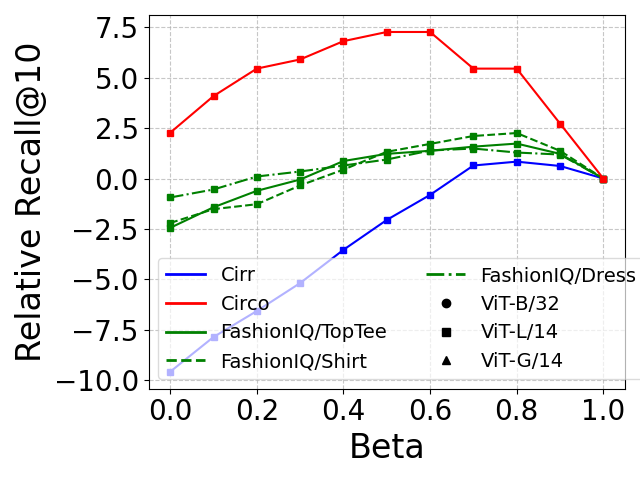}
	\includegraphics[width=0.45\linewidth]{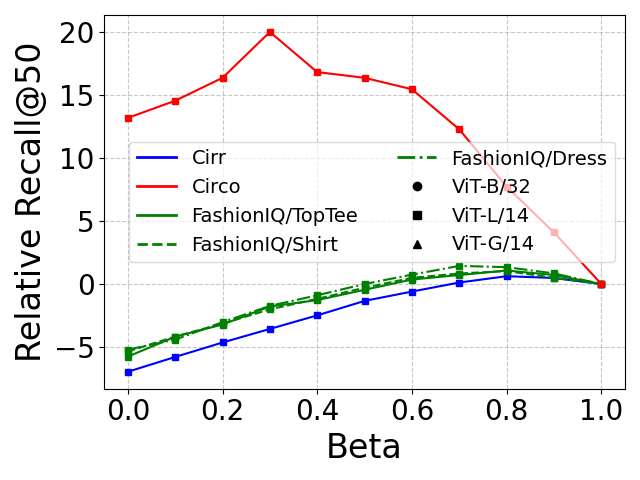}
	\\
	\includegraphics[width=0.45\linewidth]{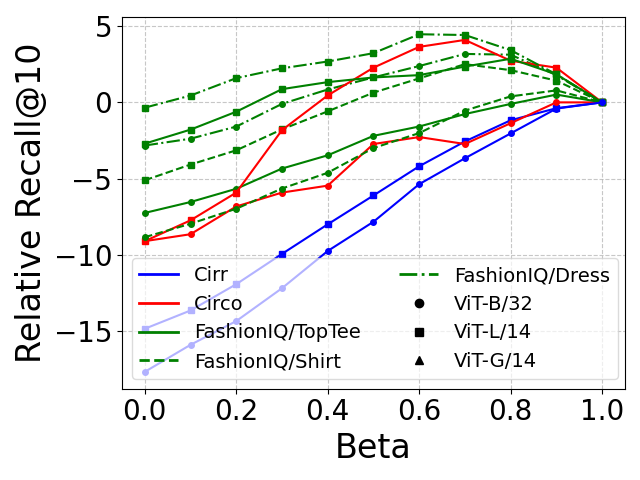}
	\includegraphics[width=0.45\linewidth]{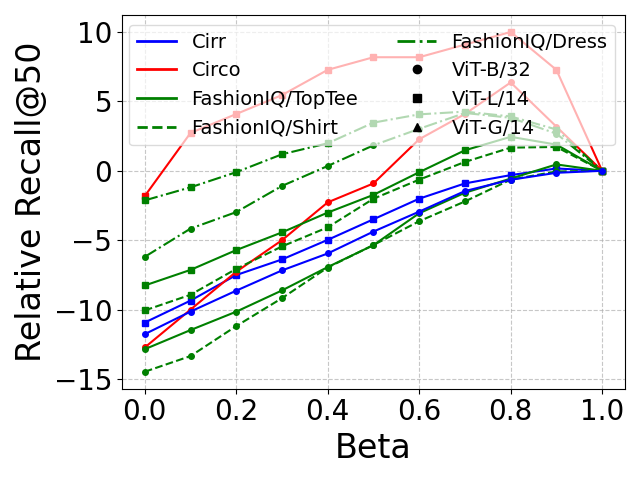}
	\caption{PDV-F: Impact of $\beta$ scaling on Recall@10 (left) and Recall@50 (right) performance. Results shown for three baseline methods: CIReVL (top), Pic2Word (middle) and SEARLE (bottom).}
	\label{fig:residual_fusion}
\end{figure*}

The PDV algorithm is given in Algorithm \ref{algo:pdv}, and the code is shown in Figure \ref{fig:pdv-code}. The implementation of PDV is very intuitive, and it could be easily integrated with any ZS-CIR approaches.

\begin{algorithm}[!th]
	\footnotesize
	\caption{Calculate PDV Features}
	\label{algo:pdv}
	\begin{algorithmic}[1]
		\Function{CalculatePDVFeatures}{$\mathbf{f}_{\text{text}}$, $\mathbf{f}_{\text{text\_composed}}$, $\mathbf{f}_{\text{image}}$, $\alpha_i$, $\alpha_t$, $\beta$}
		\State $\mathbf{f}_{\text{text}} \gets \text{normalize}(\mathbf{f}_{\text{text}})$
		\State $\mathbf{f}_{\text{text\_composed}} \gets \text{normalize}(\mathbf{f}_{\text{text\_composed}})$
		\State $\mathbf{f}_{\text{image}} \gets \text{normalize}(\mathbf{f}_{\text{image}})$
		
		\State $\mathbf{pdv} \gets \mathbf{f}_{\text{text\_composed}} - \mathbf{f}_{\text{text}}$
		
		\State $\mathbf{f}_{\text{PDVI}} \gets \mathbf{f}_{\text{image}} + \alpha_i \cdot \mathbf{pdv}$
		\State $\mathbf{f}_{\text{PDVT}} \gets \mathbf{f}_{\text{text}} + \alpha_t \cdot \mathbf{pdv}$
		
		\State $\mathbf{f}_{\text{PDVF}} \gets (1 - \beta) \cdot \mathbf{f}_{\text{PDVI}} + \beta \cdot \mathbf{f}_{\text{PDVT}}$
		
		\State \Return $\text{normalize}(\mathbf{f}_{\text{PDVF}})$
		\EndFunction
	\end{algorithmic}
\end{algorithm}

\begin{table*}[!th]
	\centering
	\small
	\begin{tabular}{l| l |c|cccccccc}
		\hline
		Fashion-IQ & & &\multicolumn{2}{c}{Shirt} & \multicolumn{2}{c}{Dress} & \multicolumn{2}{c}{Toptee} & \multicolumn{2}{c}{Average} \\ \hline
		Backbone & Method & $\alpha_{I}$ & R@10 & R@50 & R@10 & R@50 & R@10 & R@50 & R@10 & R@50 \\
		\hline
		\multirow{5}{*}{ViT-B/32} & Image-only \textdagger & -& 6.92 & 14.23 & 4.46 & 12.19 & 6.32 & 13.77 & 5.90 & 13.37 \\
		& Text-only \textdagger & - & 19.87 & 34.99 & 15.42 & 35.05 & 20.81 & 40.49 & 18.70 & 36.84 \\
		& Image + Text \textdagger & - & 13.44 & 26.25 & 13.83 & 30.88 & 17.08 & 31.67 & 14.78 & 29.60 \\
		& SEARLE + \textbf{PDV-I} & 2 & 18.25 & 31.84 & 18.49 & 39.17 & 21.32 & 37.74 & 19.35 &  36.25\\
		& CIReVL + \textbf{PDV-I} & 2 & 28.95 & 45.88 & 29.00 & 49.13 & 34.22 & 56.09 & 30.72 & 50.37 \\
		\hline
	\end{tabular}
	\caption{PDV-I performance on FashionIQ val datasets. \textdagger~denotes that numbers are taken from the original paper.}
	\label{tab:fashion_iq_results_pdv-i}
\end{table*}

\begin{table*}[!th]
	\centering
	\footnotesize
	\setlength{\tabcolsep}{4pt}
	\begin{tabular}{l|l|c|cccc|cccc|ccc}
		\hline
		\multicolumn{2}{c|}{\textbf{Dataset}} & & \multicolumn{4}{c|}{\textbf{CIRCO}} & \multicolumn{7}{c}{\textbf{CIRR}} \\
		\hline
		& Metric & & \multicolumn{4}{c|}{mAP@k} & \multicolumn{4}{c|}{Recall@k} & \multicolumn{3}{c}{$R_s$@k} \\
		\cline{2-14}
		Arch & Method & $\alpha_I$ & k=5 & k=10 & k=25 & k=50 & k=1 & k=5 & k=10 & k=50 & k=1 & k=2 & k=3 \\
		\hline
		\multirow{6}{*}{ViT-B/32} 
		& Image-only \textdagger & - & 1.34 & 1.60 & 2.12 & 2.41 & 6.89 & 22.99 & 33.68 & 59.23 & 21.04 & 41.04 & 60.31 \\
		& Text-only \textdagger & - & 2.56 & 2.67 & 2.98 & 3.18 & 21.81 & 45.22 & 57.42 & 81.01 & 62.24 & 81.13 & 90.70 \\
		& Image + Text \textdagger & - & 2.65 & 3.25 & 4.14 & 4.54 & 11.71 & 35.06 & 48.94 & 77.49 & 32.77 & 56.89 & 74.96 \\
		& SEARLE + \textbf{PDV-I} & 1.5 & 4.77 & 5.23  & 6.31 & 6.82 & 16.65 & 42.53 & 55.16 & 81.42 & 44.68 & 67.78 & 82.94\\
		& CIReVL + \textbf{PDV-I} & 2.0 & \textbf{10.29 }& \textbf{10.80} & \textbf{12.23} & \textbf{12.93} & \textbf{27.18} & \textbf{56.53} & \textbf{67.76} & \textbf{87.64} & \textbf{59.81} & \textbf{79.59} & \textbf{90.15}\\
		& LDRE + \textbf{PDV-I} & 2.0 & \underline{8.00} & \underline{8.88} & \underline{10.06} & \underline{10.72} & \underline{23.37} & \underline{51.21} & \underline{63.69} & \underline{85.57} & \underline{55.57} & \underline{76.63} & \underline{88.15}\\
		\hline
	\end{tabular}
	\caption{PDV-I performance on CIRCO and CIRR test datasets. Note that the image-only approach utilizes the visual embedding of the reference image, whereas the text-only approach employs the text embedding of the prompt. \textbf{Bold} = best results, \underline{underline} = second best. †Numbers from original paper.}
	\label{tab:circo_cirr_results_pdv-I}
\end{table*}

\end{appendices}

\clearpage
{\small
\bibliographystyle{ieee_fullname}
\bibliography{egbib}
}

\end{document}